\documentclass[twoside,twocolumn]{article}
\usepackage{microtype} 
\usepackage{graphicx}
\usepackage{booktabs} 
\usepackage{tikz}
\usetikzlibrary{calc,positioning}
\usetikzlibrary{shapes.geometric}  
\usepackage{rotating}
\usetikzlibrary{positioning,arrows.meta,calc}
\usepackage{dsfont}
\usepackage{fancyhdr}
\usepackage[T1]{fontenc}
\usepackage{microtype}
\usepackage[english]{babel}
\usepackage[hmarginratio=1:1,top=25mm, left=20mm, columnsep=10pt]{geometry} 
\usepackage[hang, small,labelfont=bf,up,textfont=up,up]{caption} 
\usepackage{abstract}

\usepackage{titling}
\usepackage{url}
\usepackage{mathtools}
\usepackage{amsmath}
\usepackage{amsthm}
\usepackage{amssymb}
\usepackage{natbib}
\usepackage{verbatim}
\usepackage{caption}
\usepackage{enumitem}
\usepackage{wrapfig}\usepackage{adjustbox}
\usepackage{bm}
\usepackage{makecell,multirow,booktabs}
\usepackage{xcolor,colortbl}
\usepackage{tabularx}
\usepackage{collcell}
\usepackage{multibib}
\usepackage[titletoc,title]{appendix}

\newcites{sec}{Additional References}

\usepackage[raggedright, compact, small]{titlesec}
\titlespacing*{\section}
{0pt}{2.5ex plus 1ex minus .2ex}{2.3ex plus .2ex}
\titlespacing*{\subsection}
{0pt}{2.5ex plus 1ex minus .2ex}{2.3ex plus .2ex}
\usepackage{anyfontsize}

\usepackage{blindtext}

\usepackage{times}
\usepackage[utf8]{inputenc} 
\usepackage[T1]{fontenc}    
\usepackage{url}            
\usepackage{booktabs}       
\usepackage{nicefrac}       
\usepackage{microtype}      
\usepackage{graphicx,color}

\usepackage{enumitem}

\usepackage{amsfonts}       
\usepackage{amsmath}       
\usepackage{amssymb}

\newcommand{\fig}[1]{Fig.~\ref{#1}}    
\newcommand{\tab}[1]{Table~\ref{#1}}

\newcommand{\eqn}[1]{Eq.~\ref{#1}} 

\usepackage{xspace}
\makeatletter
\DeclareRobustCommand\onedot{\futurelet\@let@token\@onedot}
\def\@onedot{\ifx\@let@token.\else.\null\fi\xspace}
\def\eg{e.g\onedot}
\def\ie{i.e\onedot}

\def\wrt{w.r.t\onedot}
\makeatother

\newcommand{\Real}{\ensuremath{\mathbb R}}        

\newtheorem{definition}{Definition}

\usepackage{xr}
\makeatletter
\newcommand*{\addFileDependency}[1]{
  \typeout{(#1)}
  \@addtofilelist{#1}
  \IfFileExists{#1}{}{\typeout{No file #1.}}
}
\makeatother

\usepackage{xcolor}
\definecolor{ourblue}{rgb}{0.368,0.507,0.71}
\definecolor{ourorange}{rgb}{0.881,0.611,0.142}
\definecolor{ourgreen}{rgb}{0.56,0.692,0.195}
\definecolor{ourred}{rgb}{0.923,0.386,0.209}
\definecolor{ourviolet}{rgb}{0.528,0.471,0.701}
\definecolor{ourbrown}{rgb}{0.772,0.432,0.102}
\definecolor{ourlightblue}{rgb}{0.364,0.619,0.782}
\definecolor{ourdarkgreen}{rgb}{0.572,0.586,0.}

\definecolor{ourcyan2}{rgb}{0.125,0.722,0.804}
\definecolor{ourred2}{rgb}{0.863,0.184,0.047}
\definecolor{ouryellow2}{cmyk}{0,0.16,1.0,0.07}
\definecolor{ourviolet2}{cmyk}{0.55,0.56,0,0.47}
\definecolor{ourorange2}{cmyk}{0,0.46,0.89,0.11}

\definecolor{ourcyan3}{RGB}{6, 127, 147}
\definecolor{ourred3}{RGB}{197,38,7}
\definecolor{ouryellow3}{RGB}{214,181,0}
\definecolor{ourviolet3}{RGB}{48,45,133}
\definecolor{ourorange3}{RGB}{215,111,14}

\graphicspath{{figures/}}
\usepackage{mathtools}
\usepackage{colortbl}

\usepackage[disable]{todonotes}

\newcommand{\method}{NAP}
\newcommand{\drac}{DrAC}

\newcommand{\dgM}{\mathcal{\ddot M}}

\newcommand{\Crash}{\textsc{Crash Jewel Hunt}}
\newcommand{\env}[1]{\textsc{#1}}

\usepackage{caption, subcaption}
\usepackage{algorithm}

\usepackage[noend]{algpseudocode}
\DeclareCaptionSubType*{algorithm}

\algrenewcommand\algorithmicindent{1em}
\algrenewcommand{\algorithmiccomment}[1]{%
\bgroup\hskip2em\textcolor{ourred2}{//~\textsl{#1}}\egroup}
\algrenewcommand{\Return}{\State\textbf{return}\ }
\algnewcommand{\Save}{\State\textbf{save}\ }
\algnewcommand{\Load}{\State\textbf{load}\ }

\newcommand{\fullsizeimg}[1]{\includegraphics[width=\linewidth]{#1}}

\usepackage{amsmath,amsfonts,bm}

\def\1{\bm{1}}

\def\gA{{\mathcal{A}}}

\def\gM{{\mathcal{M}}}

\def\gS{{\mathcal{S}}}

\def\gY{{\mathcal{Y}}}

\newcommand{\E}{\mathbb{E}}

\DeclareMathOperator*{\argmin}{arg\,min}

\usepackage[colorlinks=true, linkcolor=ourred3, citecolor=ourcyan3]{hyperref}

\setcitestyle{numbers,square}
\DeclareMathOperator{\JExp}{\mathop\E_{\tau_{t:t+T} \sim \eta({\pi^*})}}

\newcommand{\papertitle}{Neuro-algorithmic Policies enable Fast Combinatorial Generalization}
\newcommand{\paperrunningtitle}{Neuro-algorithmic Policies enable Fast Combinatorial Generalization}
\newcommand{\titlefont}{\fontsize{16}{20}\selectfont}

\title{\titlefont \textbf{\papertitle}}
\renewcommand{\headrulewidth}{1pt}
\renewcommand{\footrulewidth}{0pt}
\fancyheadoffset[RE,LO]{0cm}
\author{
{Marin Vlastelica, Michal Rol\'inek, Georg Martius} \\[1ex]
\normalsize Max Planck Institute for Intelligent Systems\\
\normalsize {\tt\small \{mvlastelica, mrolinek, gmartius\}@tue.mpg.de} 
}
\date{}
\renewcommand{\maketitlehookd}{

\begin{center}
  \fullsizeimg{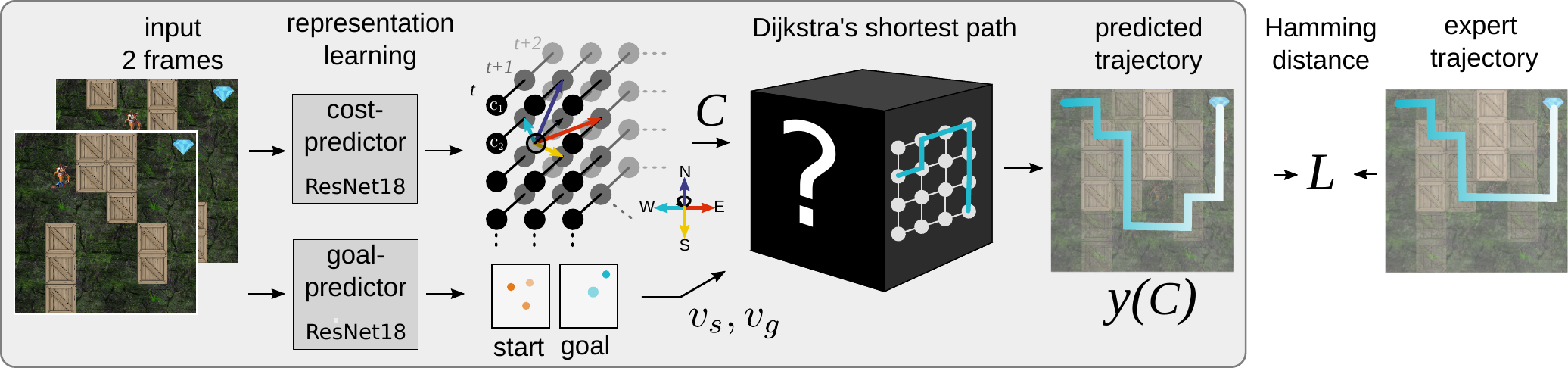}
  \captionof{figure}{Architecture of the neuro-algorithmic policy. Two subsequent frames are processed by two simplified ResNet18s: the cost-predictor outputs a tensor (width $\times$ height $\times$ time) of vertex costs $C^t(v)$ and the goal-predictor outputs heatmaps for start and goal. The time-dependent shortest path solver finds the shortest path to the goal. Hamming distance between the proposed and expert trajectory is used as loss for training.}
  \label{fig:pipeline}
 \end{center}

}

\begin{document}

\newcolumntype{Y}{>{\centering\arraybackslash}X}
\renewcommand\tabularxcolumn[1]{m{#1}}

\maketitle

\newcommand{\fix}{\marginpar{FIX}}
\newcommand{\new}{\marginpar{NEW}}

\begin{abstract}
Although model-based and model-free approa\-ches to learning the control of systems have achieved impressive results on standard benchmarks, generalization to task variations is still lacking.
Recent results suggest that generalization for standard architectures improves only after obtaining exhaustive amounts of data.
We give evidence that generalization capabilities are in many cases bottlenecked by the inability to generalize on the combinatorial aspects of the problem.
Furthermore, we show that for a certain subclass of the MDP framework, this can be alleviated by neuro-algorithmic architectures.

Many control problems require long-term planning that is hard to solve generically with
neural networks alone.
We introduce a neuro-algorithmic policy architecture consisting of a neural network and an embedded time-dependant shortest path solver. 
These policies can be trained end-to-end by blackbox differentiation.
We show that this type of architecture generalizes well to unseen variations in the environment already after seeing a few examples.
\end{abstract}
\vfill
\goodbreak


\section{Introduction}
One of the central topics in machine learning research is learning control policies for autonomous agents.
Many different problem settings exist within this area.
On one end of the spectrum are imitation learning approaches, where prior expert data is available and the problem becomes a supervised learning problem.
On the other end lie approaches that require interaction with the environment to obtain data for policy extraction, posing the problem of exploration.
Most Reinforcement Learning (RL) algorithms fall into the latter category.
In this work, we concern ourselves primarily with the setting where limited expert data is available, and a policy needs to be extracted by imitation learning.


Independently of how a policy is extracted, a central question of interest is: how well will it generalize to variations in the environment and the task?
Recent studies have shown that standard deep RL methods require exhaustive amounts of exposure to environmental variability before starting to generalize~\citep{CobbeEtAL2019:procgen}.

There exist several approaches addressing the problem of generalization in control.
One option is to employ model-based approaches that learn a transition model from data and use planning
algorithms at runtime or to improve value-learning.
This has been argued to be the best strategy in the presence of an accurate model and sufficient computation time \citep{daw2005uncertainty}.
Furthermore, one can use the transition model alongside a reward model to generate offline data to improve value function learning \citep{sutton1991dyna, janner2019trust}.
However, learning a precise transition model is often harder than learning a policy.
The transition model often has a much larger dimensionality than the policy since it needs to model aspects of the environmental dynamics that are perhaps irrelevant for the task.
This is particularly true for learning in problems with high-dimensional inputs, such as raw images.
In order to alleviate this problem, learning specialized or partial models has shown to be a viable alternative, \eg in MuZero~\citep{schrittwieser2019mastering}.

We propose to use recent advances in making combinatorial algorithms differentiable in a blackbox fashion\citep{VlastelicaEtal2020} to train neuro-algorithmic policies with embedded planners end-to-end.
More specifically, we use a time-dependent shortest path (TDSP) planner acting on a temporally evolving graph generated by a deep network from the inputs.
By learning the time-evolving costs of the graph our method builds a specific model of the system that is sufficient for planning.
We demonstrate the effectiveness of this approach in an offline imitation learning setting where
 a few expert trajectories are provided.
Due to the combinatorial generalization capabilities of planners, our learned policy is able
 to generalize to new variations in the environment out of the box and needs orders of magnitude fewer samples than
 naive learners.
Using neuro-algorithmic architectures facilitates generalization by shifting the combinatorial aspect of the problem to efficient algorithms, while using neural networks to extract a good representation for the problem at hand.
They have the potential to endow artificial agents with the main component of intelligence, the ability to reason.


Our contributions can be summarized as follows:
{\bf(i)} We show that combinatorial inductive biases implemented through neuro-algorithmic policies can be used to tackle the generalization problem in reinforcement learning.
{\bf(ii)}    We show that architectures embedding TDSP solvers are applicable beyond goal-reaching environments.
{\bf(iii)}    We demonstrate learning neuro-algorithmic policies in dynamic game environments from images.


\section{Related Work}

\paragraph{Planning} There exist multiple lines of work aiming to improve classical planning algorithms such as improving sampling strategies of Rapidly-exploring Random Trees   \citep{gammell2014informed, burget2016bi, kuo2018deep}.
Similarly, along this direction, \citet{kumar2019LEGOLE} propose a conditional VAE architecture for sampling candidate waypoints.
Orthogonal to this are approaches that learn representations such that planning is applicable in the latent space. \citet{Hafner2019:PlaNet} employ a latent multistep transition model.
\citet{savinov2018semi} propose a semi-parametric method for mapping observations to graph nodes and then applying a shortest path algorithm. \citet{asai2017classical, asai2019towards} use an autoencoder architecture to learn a discrete transition model suitable for classical planning algorithms.
\citet{LiTorralba2020:CompKoopman} learn compositional Koopman operators with graph neural networks mapping to a linear dynamics latent space, which allows for fast planning.
\citet{chen2018optimal, amos2017input} perform efficient planning by using a convex model formulation and convex optimization.
Alternatively, the replay buffer can be used as a nonparametric model to select waypoints \citep{eysenbach2019search} or in an MPC fashion \citep{blundell2016model}.
None of these methods perform differentiation through the planning algorithm to learn better latent  representations.

\paragraph{Differentiation through planning} Embedding differentiable planners has been proposed in previous works, \eg in the continuous case with CEM \citep{br2019differentiable,bharadhwaj2020model}.
\citet{wu2020scalable} use a (differentiable) recurrent neural network as a planner.
\citet{tamar2016value} use a differentiable approximation of the value iteration algorithm to embed it in a neural network.
\citet{Silver2017:end2endlearningandplanning} differentiate through a few steps of value prediction in a learned MDP to match the externally observed rewards.
\citet{Srinivas2018:UPN} use a differentiable forward dynamics model in latent space.
\citet{karkus2019differentiable} suggest a neural network architecture embedding MDP and POMDP solvers and during the backward pass, they substitute the algorithms by learned approximations.
In comparison, we do not perform any relaxation or approximation of the planner itself and we learn interpretable time-dependent costs of the latent planning graph based on expert demonstrations by differentiating through the planner.
Similarly to our work, \citet{yonetani2020path} embed an A$^*$ algorithm into a neural network, but in comparison, their method does not operate with time-dependent costs, subgoal selection and does not provide a policy for closed-loop control.


\paragraph{Inverse reinforcement learning and imitation learning} Uncovering the expert's objective function from demonstrations has been a central topic in reinforcement learning \citep{ng2000algorithms}.
Our method is connected to inverse reinforcement learning in the sense that we learn the objective
function that the expert optimizes to extract an optimal policy, also called apprenticeship learning \citep{abbeel2004apprenticeship, neu2012apprenticeship, aghasadeghi2011maximum}.
What separates our approach is that the inferred costs are inherently part of the learned neuro-algorithmic policy in conjunction with the applied planner on the costs.

 Our method is an offline imitation learning method, but since we propose an end-to-end trainable policy, it is naturally extendable to the online case with a method such as DAgger~\citep{ross2011reduction} or other online reinforcement learning methods augmented with expert datasets~\citep{ reddy2019sqil,ho2016generative}.

\paragraph{Offline model-based reinforcement learning} Model-based methods have shown promise by facilitating better generalization \citep{janner2019trust}.
These approaches fall into two camps: using models to extract a policy in a Dyna-style approach \citep{sutton1991dyna, janner2019trust, sutton2008dyna, yao2009multi, kaiser2019model}, or incorporating the model in a planning loop, \ie model-predictive control \citep{finn2017deep, racaniere2017imagination, oh2017value, silver2017predictron,pinneri2021:strong-policies}.
In this work, we consider the latter case where an implicit transition model is ``hidden'' within the predicted time-dependent costs.

\paragraph{Combinatorial algorithms in end-to-end trainable networks}
We suggest a hybrid policy consisting of a neural network and an accompanying expert (shortest path) discrete solver that is trainable end-to-end.
Incorporating expert discrete solvers into end-to-end trainable architectures is a topic with exciting recent developments.
For the simpler setup of comparing to ground-truth values on the solver output, numerous frameworks have been suggested such as the ``predict-and-optimize'' framework and its variants \citep{predict-optimize, predict-optimize-ranking, predict-and-optimize-comb}.
Moreover, specializations for concrete cases such as sparse structured inference~\citep{niculae2018sparsemap},
logical satisfiability~\citep{wang2019satnet},
submodular optimization~\citep{djolonga2017differentiable}, or
mixed integer programming \citep{ferber2020mipaal} have been proposed.

We are interested in the harder case of providing an \textit{entirely hybrid} architecture which may use the solver at intermediate levels and is trainable end-to-end.
For this case,  two approaches have recently emerged \citep{VlastelicaEtal2020, berthet2020learning}. 
\citet{VlastelicaEtal2020} introduce an efficient implicit piecewise linear interpolation scheme, while \citet{berthet2020learning} introduce an estimation of a Gaussian smoothing of the piecewise constant function.
The approach from \citet{VlastelicaEtal2020} is especially appealing, since it allows for uses in which the solver is the computational bottleneck.
By formulating the control problem as a time-dependent shortest path problem (TDSP), we show that the framework of \citet{VlastelicaEtal2020} is applicable in specific control settings.

\section{Markov Decision Processes and Shortest Paths}
\label{sec:mdp}
We consider the problem formulation in the context of  a Markov Decision Process (MDP).
We concern ourselves with policies that solve a time-dependent shortest path problem on a latent graph representation related to that MDP.
We start with a class of MDPs that can be directly mapped to a shortest path problem and construct this mapping.
Then we consider conditions that allow for a reduced latent graph
where the optimal policy follows a time-dependent shortest path problem.

First, we consider discrete MDPs with deterministic transitions.
In addition, we follow a goal-conditioned setting~\citep{schaul2015:UVFA}.
This is used in sequential decision making problems where a specific terminal state has to be reached.
\begin{definition}
\label{def:gcmdp}
A goal-conditioned Markov Decision Process (gcMDP) $\gM$ is defined by the tuple  ($\gS$, $\gA$, $p$, $g$, $r$), where $\gS$ is the state space, $\gA$ the action space, $p(s'\mid a, s)$ the probability of making the transition $s \in \gS \rightarrow s' \in \gS$ when taking the action $a \in \gA$, $g \in \gS$ is the goal, $r(s, a, s', g)$ the reward obtained when transitioning from state $s$ to $s'$ while taking action $a$ and aiming for goal $g$.
\end{definition}
In episodic reinforcement learning, the objective is to find a policy that maximizes the return $G = \sum_{t=0}^T r_t$ of such a process. In gcMDPs, the reward is such that the maximal return can be achieved by reaching the goal state $g$, which is also the terminal state.

\begin{definition}
\label{def:dgcmdp}
A discrete and deterministic goal-conditioned Markov Decision Process (ddgcMDP) $\dgM$ is a gcMDP with discrete and finite state space $\gS$ and action space $\gA$ with deterministic transitions, \ie  $p(s'\mid a, s)$ is one-hot.  \end{definition}

Let us consider the following graph representation $(G,v_s,v_g)$:
a weighted graph $G = (V, E, C)$ together with start vertex $v_s$ and goal vertex $v_g$,
where $V$ is the vertex set, $E\subset (V \times V)$ is the edge set, and $C\in \Real_+^{|E|}$ is the cost matrix with positive entries. We write $C(e)$ for the cost of edge $e$.
In addition, we consider an inverse model $\psi\colon E \to \mathcal A$, associating an edge to an action.

A direct translation of a ddgcMDP to a graph is given by a bijective correspondence $\phi\colon \mathcal S \to V$ between states and vertices,
for each possible transition $(s,a,s')$ there is an edge $e=(\phi(s),\phi(s'))$ with  $\psi(e) = a$, and the goal vertex is $v_g=\phi(g)$.
The cost matrix $C$ takes the role of the reward function with $C(\phi(s),\phi(s')) = c_{\mathrm{max}} - r(s,a,s',g)$ where $c_{\mathrm{max}}$ is an upper bound on the reward ensuring positive cost values.
Due to the deterministic nature of the ddgcMDP, the optimal policy yields a path with maximal return (sum of rewards) from start $s_0$ to goal $g$, which now coincides with the shortest path according to $C$ by definition.

\subsection{Factorized MDPs and Time-dependent Shortest Path}
In many interesting environments, the state space is exponentially large, \eg due to  independent dynamics of entities.
As an illustrative example, consider the game environment shown in \fig{fig:pipeline} and \ref{fig:crash} -- an agent with moving obstacles.
For such environments, the corresponding graph would become intractably large.
We now define conditions under which the size of the graph can be drastically reduced, lwithout losing the property that its shortest path solves the MDP.
To this end, we assume the state space can be factorized as
 $\gS = \gS_A \times \gS_E$, where $\gS_A$
 is affected only by the actions and $\gS_E$ by the environment dynamics independent of the actions. We write the decomposition of each state as $s=(s^a,s^e)$.
The mapping from state space to graph vertices $\phi\colon \mathcal S \to V$ is bijective only \wrt $\gS_A$ and ignores $\gS_E$, 
\ie~$\forall s^a \in \gS_A \; \exists\, v \in V\colon \phi(s^a, s^e) = v \,, \forall \, s^e \in \gS_E$. For brevity, we write $\phi(s^a)$.
For instance, using $\gS_A$ as the agent's positions on a discrete grid results in
just as many vertices as there are positions of the agent.

Next, we show how a solution to this factorized ddgcMDP can be found by solving a  time-dependent shortest path (TDSP) problem on the reduced graph.
The cost matrix $C$ is now a time-dependent quantity, \ie
$C\in \Real_+^{{T \times |E|}}$ that assigns every edge $e$ a positive cost value for each time $t \in \{1,2,\dots, T\}$ of the planning horizon/episode. To ease the notation, we write $C^t(e)$ for the cost of edge $e$ at time $t$.
The TDSP problem is defined as reaching
the goal vertex $v_g$ within at most $T$ steps when starting at time step $1$ in the start vertex $v_s$.

Two situations need to be modeled by the cost:%
\footnote{Note that learning the \emph{exact} costs is not necessary, since we are only interested in optimal trajectories.}
a) the environment dynamics can make a transition along an unavailable edge $e$ (for instance, an obstacle is moving in the way),
 then the cost should be infinite: $C^t(e)=\infty$, and
b) the environment dynamics change the reward, thus for all other edges we have
\begin{align}
     C^t(\phi(s^a),\phi(s'{^a})) = c_{\mathrm{max}} - r((s^a,s_t^e), a, (s'{^a},s_t^e))
     \label{eqn:cost:construction}
\end{align}
where $a=\psi(\phi(s^a),\phi(s'{^a}))$.
Again, with this construction, the time-dependent shortest path solution coincides with the optimal policy of the specific ddgcMDP.
In fact, we can also deal with stochastic environment dynamics as long as the action's effect on the state stays deterministic.
This changes the reward term in \eqn{eqn:cost:construction} to an expectation over environment dynamics:
\begin{align*}
 \underset{s_t^e}{\mathbb{E}}[ r((s^a,s_t^e), a, (s'{^a},s_t^e))].
\end{align*}
In our architecture, described below, the policy generates a latent graph
at every observation/state and solves a ddgcMDP to optimality every time step
following a model predictive control approach using
receding horizon planning.


\section{Shortest Path Algorithm and its Differentiation}
We aim to predict the cost matrix $C$ via a deep neural network and train the entire system end-to-end via gradient descent on expert trajectories, as illustrated in \fig{fig:pipeline}.
We will employ an efficient implementation of Dijkstra's algorithm for computing the shortest path. For differentiation, we rely on the framework for blackbox differentiation of combinatorial solvers~\cite{VlastelicaEtal2020}.

\paragraph{Time-dependent shortest path with vertex costs.}
In our neuro-algorithmic policy, we use the \textsc{time-dependent-shortest-path} (TDSP) formulation
 based on vertex-costs instead of edge-costs, due to the reduction in cost matrix size by a factor of $|\gA|$.
The TDSP solver has as input the graph $G(V,E,C)$ with time-dependent cost matrix
$C\in\Real_+^{{T \times |V|}}$ and a pair of vertices $v_s, v_g \in V$ (start and goal).
We write $C^t(v)$ for the cost of vertex $v$ at time $t$.

This version of the shortest path problem can be solved by executing the Dijkstra's shortest path algorithm\footnote{Even though the classical formulation of Dijkstra's algorithm is edge-based, all of its properties hold true also in this vertex-based formulation.} \cite{Dijkstra} on an augmented graph. In particular, we set
\begin{align*}
V^* &= \{(v, t)\colon v \in V, t \in [1,T]\}\quad \\
E^* &= \{((v_1, t), (v_2, t\!+\!1))\colon (v_1, v_2) \in E^{\circlearrowleft}, t \in [1,T\!-\!1]\},
\end{align*}
where the cost of vertex $(v_i, t) \in V^*$ is simply $C^t(i)$ and $E^{\circlearrowleft}$ is the original edge set $E$ appended with all self-loops.
This allows to ``wait'' at a fixed vertex $v$ from timestep $t$ to timestep $t+1$.
In this graph, the task is to reach the vertex $(v_g, T)$ from $(v_s, 1)$ with the minimum traversal cost.


\subsection{Applicability of Blackbox Differentiation}

\begin{figure}[tb]
  \centering
  \includegraphics[width=1.0\linewidth]{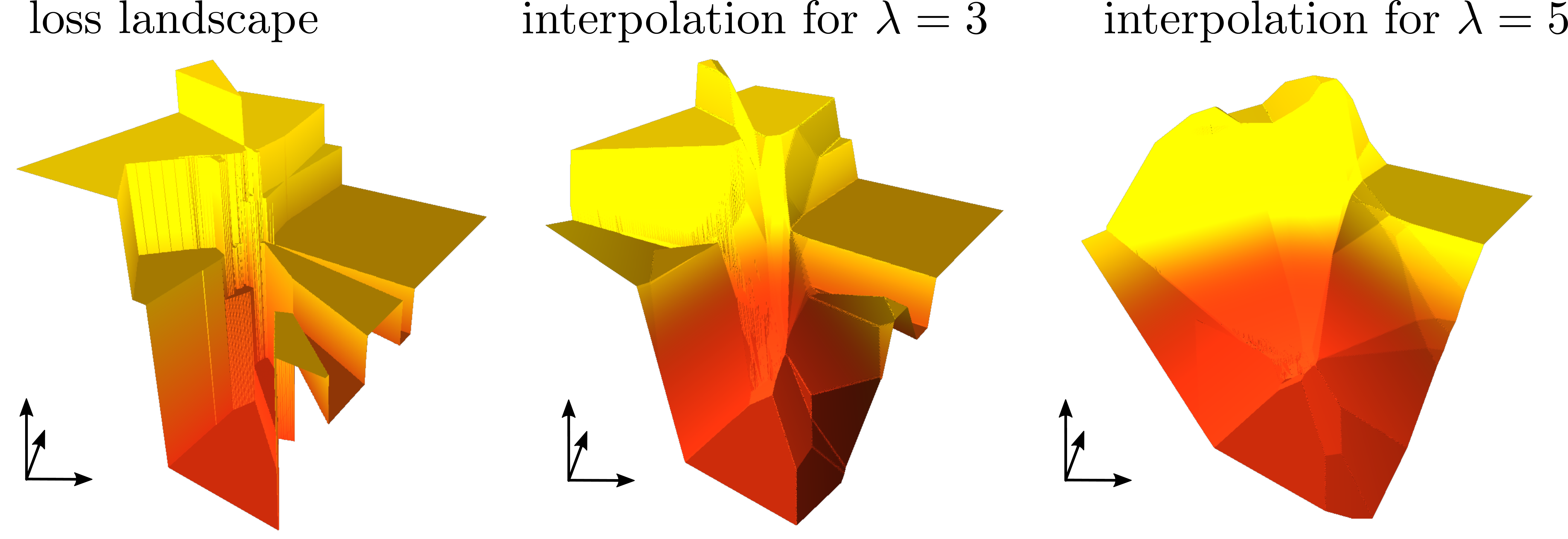}
	\caption{Differentiation of a piecewise constant loss resulting from
	incorporating a combinatorial solver. A two-dimensional section of the loss
	landscape is shown (left) along with two differentiable interpolations of
	increasing strengths (middle and right).
	}
  \label{fig:f-lambda-2d}
\end{figure}

The framework presented in \cite{VlastelicaEtal2020} turns blackbox combinatorial solvers into neural network building blocks. The provided gradient is based on a piecewise linear interpolation of the true piecewise constant (possibly linearized) loss function, see \fig{fig:f-lambda-2d}.
The \emph{exact} gradient of this linear interpolation is computed efficiently via evaluating the solver on only one additional instance (see Algo.~\ref{alg:main}).

\begin{algorithm}[tb]
\begin{algorithmic}
		\Function{ForwardPass}{$C$, $v_s$, $v_g$}
		\State
		$Y := \text{\bfseries TDSP}(C, v_s, v_s)$\hfill \Comment{Run Dijkstra's algo.}
		\Save $Y$, $C$, $v_s$, $v_e$ \hfill \Comment{Needed for backward pass}
		\Return $Y$
		\EndFunction
		\medskip
		\Function{BackwardPass}{$\nabla L(Y)$, $\lambda$}
		\Load $Y$, $C$, $v_s$, $v_e$
		\State $C_\lambda
			:= C + \lambda\nabla L(Y)$\hfill \Comment{Calculate modified costs}
		\State $Y_\lambda
			:= \text{\bfseries TDSP}(C_\lambda, v_s, v_g)$\hfill
		\Comment{Run Dijkstra's algo.}
		\Return $\frac1\lambda\bigl(Y_\lambda-Y \bigr)$
		\EndFunction
	\end{algorithmic}
\caption{Forward and backward Pass for the shortest-path algorithm}\label{alg:main}%
\end{algorithm}

To apply this differentiation scheme, the solver at hand needs to have a formulation in which it minimizes an inner-product objective (under arbitrary constraints).
To that end,
for a given graph $G = (V, E, C)$ as describe above,
 we define $Y \in \{0,1\}^{T \times |V|}$ an indicator matrix of visited vertices.
In particular, $Y_i^t = 1$ if and only if vertex $v_i$ is visited at time point $t$.
The set of such indicator matrices that correspond to valid paths in the graph $(V^*, E^*)$ from start to goal will be denoted as $\gY \coloneqq \mathrm{Adm}(G, v_s,v_g)$.
The time-dependent shortest path optimization problem can be then rewritten as
\begin{align}
\mathrm{TDSP}(C,v_s,v_g) &= \argmin_{Y \in \gY} \sum_{(i, t)} Y_i^t C_i^t,
\end{align}
where $\gY =  \mathrm{Adm}(G, v_s, v_g)$.
This is an inner-product objective and thus the theory from \cite{VlastelicaEtal2020} applies.
and allows us to learn the a cost-matrix generating network with gradient descent, as described below.

\subsection{Cost Margin}
The supervision-signal for the costs $C$ of the latent graph  is only indirectly given via the shortest path solutions.
At training time there is no incentive to make these costs robust to small misestimations.
Thus, inducing a margin on costs can be beneficial for generalization.
Similarly to \citet{rolinek2020cvpr}, where it was used in the context rank-based metric learning,
we induce a margin $\alpha$ by increasing the cost of the ground-truth path and decreasing the rest in the training stage of the algorithm:
\begin{equation} \label{eq:margin}
    c_i^t =
	\begin{cases}
		c_i^t + \frac \alpha 2
			& \text{if $(v_i, t) \in Y^*$}
			\\
		c_i^t - \frac \alpha 2
			& \text{if $(v_i, t) \notin Y^*$}
	\end{cases}
	\; \forall \, (v_i, t) \in V^*.
\end{equation}

\section{Neuro-algorithmic Policy Framework}

We propose the Neuro-algorithmic Policy (\method) framework,
which is an end-to-end trainable
deep policy architecture embedding an algorithmic component
using the afore-mentioned techniques.
Following the definitions from Sec. \ref{sec:mdp}, we concern ourselves with learning the mapping $\varphi_\theta: \mathcal S \mapsto \mathbb{R}^{|V| \times T} \times V \times V$, \ie mapping from MDP states to cost matrices and respective start and end vertices for the $\mathrm{TDSP}$ problem
\footnote{We hold the graph structure fixed, namely the set of edges $E$ and learn the costs $C$. Therefore we replace $G$ with costs $C$ in the $\mathrm{TDSP}$ solver to simplify the notation.} 
of planning horizon $T$.
This enables us to construct the policy
\begin{equation}
    \pi_\theta \coloneqq  \psi \circ \mathrm{TDSP}  \circ \varphi_\theta.
\end{equation}
The mapping $\varphi_\theta$ can be decomposed into $\varphi_\theta^c$ (\emph{cost-predictor}), $\varphi_\theta^s$ (\emph{start-predictor}), $\varphi_\theta^g$ (\emph{goal-predictor}), \ie mappings from state to costs, start vertex and goal vertex.
In practice, instead of learning $\varphi_\theta^s$ and $\varphi_\theta^g$ directly, we learn the conditional probability densities $p_\theta^s(v | s)$ and $p_\theta^g(v | s)$.

For this work, we examine the application of neuro-algorithmic policies to the imitation learning setting, where we have access to trajectories $\tau$ sampled from the expert policy distribution $\eta(\pi^*)$.
Given a fixed planning horizon $T$, the objective that we consider consists of three main parts, the latent cost term, start vertex term and goal vertex term.
The latent cost objective is defined as
\begin{equation}
J^C(\theta, T) =  \JExp \big[ \mathrm{d^H}(\mathrm{TDSP}(\varphi_\theta(\tau_t)), \phi'(\tau))\big],
\end{equation}
where $\mathrm{d^H}(\cdot, \cdot)$ denotes the Hamming distance between predicted and expert paths in the latent graph, and $\phi'\colon \mathcal{S} \mapsto V$ the mapping of the expert-visited states to latent graph vertices.
The second part of the objective is a standard cross-entropy term for the start and goal vertices that allows us to train a \emph{start-} and \emph{goal-predictor}:
\begin{align}
\begin{split}
    J^P(\theta, T) =  \JExp \big[&-\log p_\theta^s(\phi'(\tau_t) | \tau_t ) \\
               &- \log p_\theta^g(\phi'(\tau_{t+T}) | \tau_t)\big].
\end{split}
\end{align}
We assume access to $\phi'$ at training time in order to map the expert to the latent graph for calculation of $J^C$ and $J^P$.
Finally, we optimize for the sum of $J^C$ and $J^P$.


We utilize a concrete architecture consisting of two main components:
a backbone ResNet18 architecture (without the final fully connected layers (a detailed description is available in Sec.~\ref{app:architecture} in the supplementary) and the shortest path solver, see \fig{fig:pipeline}.
At each time step, the policy receives two images concatenated channel-wise from which it predicts the cost tensor $C$ for the planning horizon $T$ with the \emph{cost-predictor}, the start vertex $v_s$ and goal vertex $v_g$ with the \emph{goal-predictor}, explained below.

The policy is used in a model-predictive control fashion, \ie at execution time, we predict the plan $Y$ for horizon $T$ at each time step and execute the first action from the plan.

\subsection{Global and Local Goal Prediction}
In order to apply the solver to the learned latent graph representation,
we need to map the current state of the environment to appropriate start and goal vertices ($v_s,v_g$).
To this end, we employ a second ResNet18  similar to the \emph{cost-predictor} that approximates $p^s(v| s)$ and $p^g(v | s)$, \ie the start and goal conditional densities.

\begin{figure}[t]
  \centering
  \begin{tabular}{c@{\hspace{5mm}}c}
  (a) Frame & (b) Game logic \\
  \includegraphics[width=0.4\linewidth]{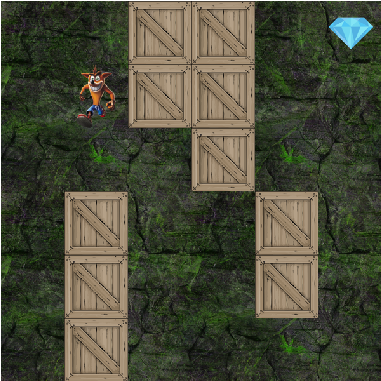} &
  \includegraphics[width=0.4\linewidth]{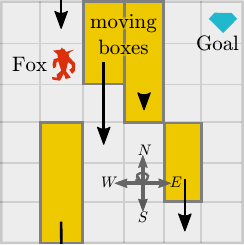} \\
  \multicolumn{2}{c}{(c) Cost prediction example}\\
  \multicolumn{2}{c}{\includegraphics[width=0.9\linewidth]{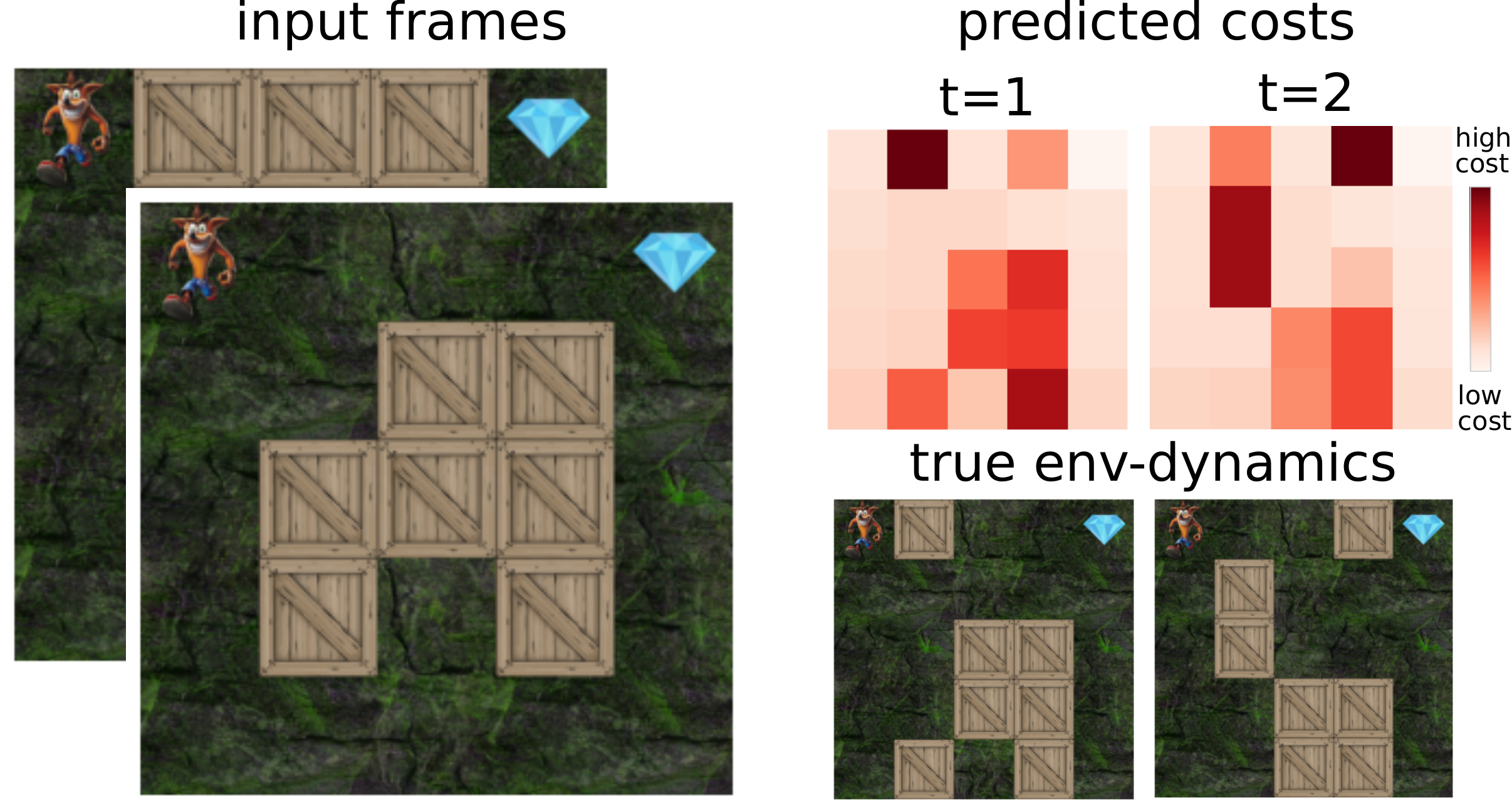}} %
  \end{tabular}
  \caption{The \Crash{} environment. 
	The goal for the fox, see (a), is to obtain the jewel in the right most column, while avoiding the moving wooden boxes (arrows in (b)).
	When the agent collides with a wooden box it instantly fails to solve the task. We observe that the predictions of the costs in (c) are highly interpretable, corresponding to (future) movements of the boxes.}
  \label{fig:crash}
\end{figure}

At training time, given the expert trajectories, we have access to the mapping $\phi'$ that maps the expert trajectory to the latent graph.
In the \emph{global} setting, the last position of the expert is the goal $v_g$, corresponding to, for instance, the jewel in \Crash{}  which is also the terminal state, see Fig. \fig{fig:crash}.

In the \emph{local} setting, we expect the end vertex to be an intermediate goal (``collect an orb''), which effectively allows for high-level planning strategies, while the low-level planning is delegated to the discrete solver.
In this case, the positively labeled supervision at time $t$ are all locations of the (expert) agent between step $t+T$ and $t+2T$.

The local setting allows to limit the complexity of our method, which grows with the planning horizon.
This is also a trade-off between the combinatorial complexity solved by the $TDSP$ solver and the goal predictor.
Ideally, the planning horizon $T$ used for the cost-prediction is long enough to capture the combinatorial intricacies of the problem at hand, such as creating detours towards the goal in the case of future dangerous states, or avoiding dead-ends in a maze.

This formulation makes our architecture akin to hierarchical methods similar to  \citet{blaes2019control,nachum2018data},
and allows for solving tasks that are not typical goal-reaching problems, such as the \textsc{Chaser} environment.

\section{Experiments}
To validate our hypothesis that embedding planners into neural network architectures leads to better generalization in control problems,
we consider several procedurally generated environments (from the ProcGen suite~\citep{CobbeEtAL2019:procgen} and \Crash{}) with considerable variation between \emph{levels}.

We compare with the following baselines: a standard behavior cloning (BC) baseline using a ResNet18 architecture trained with a cross-entropy classification loss on the same dataset as our method; the PPO algorithm as implemented in \citet{CobbeEtAL2019:procgen} and data-regularized actor-critic (DrAC) \cite{raileanu2020automatic}.
A comparison to DrAC is especially interesting, since the method claims to improve generalization by applying optimized data augmentations.
As there are multiple variations of data augmentation suggested by \citet{raileanu2020automatic}, we run all of them and select the best result as DrAC$^*$ for performance comparison to our method\footnote{We defer a more detailed description of DrAC along with performance plots to Sec. \ref{app:drac-details}}.
We also ablate the \emph{start-} and \emph{goal-predictor} and replace with the ground truth vertices, this serves as an upper baseline for \method{} which we denote with \method$^*$.
More details on the training procedure and the hyperparameters can be found in the supplementary Sec.~\ref{app:training}.

For the experimental validation, we aim to anwser the following questions: \textbf{(i)}~Can \method{} be trained to perform well in procedurally generated environments? \textbf{(ii)}~Can \method{} generalize in a low data regime, \ie after seeing only few different levels? \textbf{(iii)}~Can we also solve non-goal-reaching environments?


\newlength{\pich}
\begin{figure*}[t] \centering
        \setlength{\pich}{0.164\linewidth}
        \begin{tabular}{@{}c@{\ \ }c@{\ \ }c@{\ \ }c@{}}
             \textsc{Crash $5\times 5$} &          \textsc{Crash $5\times 10$} & \textsc{Maze} & \textsc{Leaper(Grid)} \\
             \includegraphics[height=\pich]{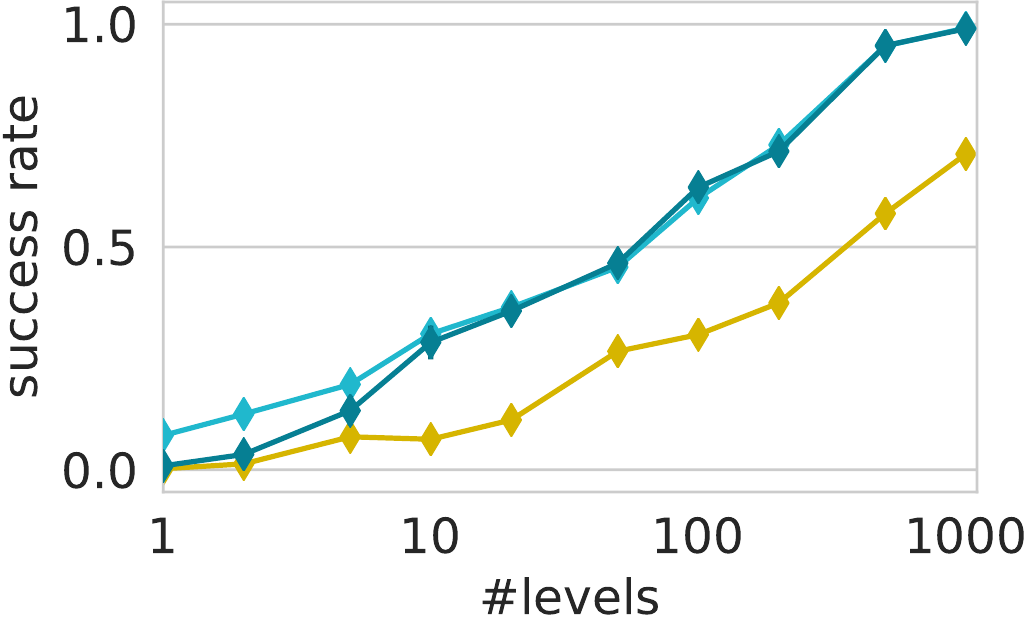}
             &
             \includegraphics[height=\pich]{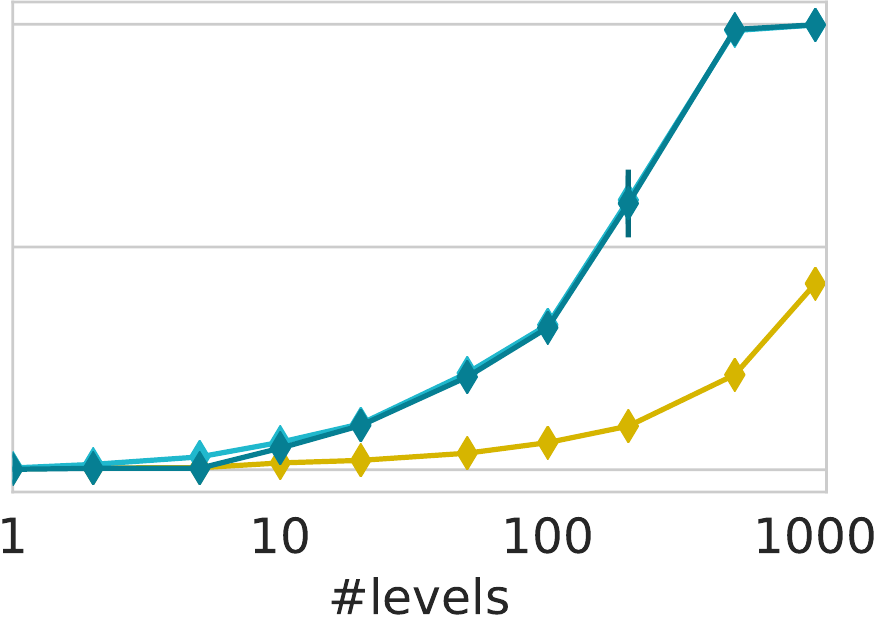}
             &
             \includegraphics[height=\pich]{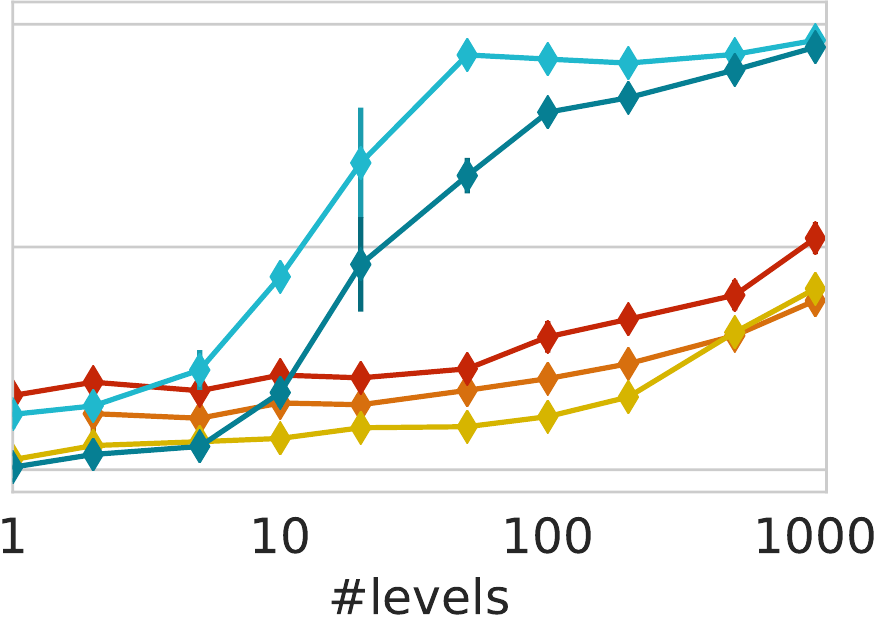}
             &
             \includegraphics[height=\pich]{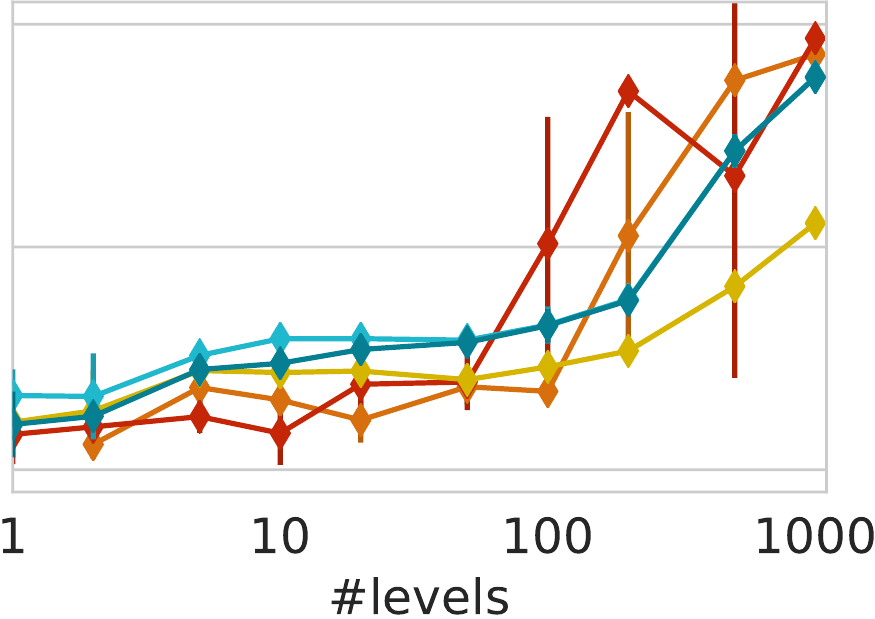} \\
             \multicolumn{4}{c}{
                 \includegraphics[width=.5\linewidth]{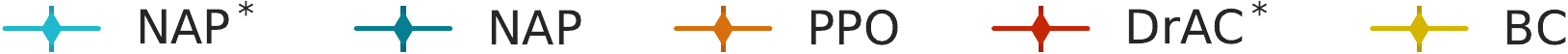}
             }
        \end{tabular}\vspace{-.7em}
        \caption{Performance on unseen test levels for a different number of training levels.
        We observe that \method{} already shows signs of generalization after only being trained on 100 levels.}

        \label{fig:performance-plots}
    \end{figure*}

\subsection{\Crash{}}

For proof of concept, we first consider an environment we constructed to test \method{}, called \Crash{} which can be seen in \fig{fig:crash}.
The environment corresponds to a grid-world of dimensions $h \times w$ where the goal is to move the agent (Fox) from an arbitrary start position in the left-most column to the goal position (jewel) arbitrarily positioned in the right-most column.
Between the agent and the goal are obstacles, wooden boxes that move downwards (with cyclic boundary conditions) with velocities that vary across levels but not within a level, see \fig{fig:crash}~(right).
At each time step, the agent can choose to move horizontally or vertically in the grid by one cell or take no action.

To make the task challenging, we sample distinct environment configurations for the training set and the test set, respectively.
More concretely, we vary the velocities, sizes and initial positions of the boxes as well as the start and goal positions.

\subsection{ProcGen Benchmark}

In addition to the jewel hunt environment, we evaluate our method on the hard version \env{Maze}, \env{Leaper} and \env{Chaser} environments from the ProcGen suite \citep{CobbeEtAL2019:procgen}.
We have chosen these environments because their structure adheres to our assumptions.
For \env{Leaper}, we modified the environment such that grid-world dynamics apply (\env{Leaper(Grid)}).

The \env{Maze} and the \env{Leaper(Grid)}  tasks have a static goal whose position only varies across levels, whereas the \env{Chaser} requires collection of all orbs without contact with the spiders, so the local goals need to be inferred on the fly. The \env{Chaser} environment is also \textbf{particularly challenging} as even the expert episodes require on average 150 steps, most of which carry the risk of dying. For this reason, we used three human expert trajectories per level.

\subsection{Results}

We train our method (\method{}) and the imitation learning baseline until saturation on a training set, resulting in virtually $100\%$ success rate when evaluating on train configurations in the environment.
For the PPO baseline we use the code from~\cite{CobbeEtAL2019:procgen} and provide also two subsequent frames and 200M time steps for training.
For the DrAC baselines we use the code from~\citet{raileanu2020automatic}.
For our method, we also report performance of a version with access to the true start and end-point prediction (\method$^*$), with the exception of the \env{Chaser} where true goals are not well-defined.

\begin{figure}[h]
    \centering
    (a) Performance on \textsc{Chaser}\\
    \begin{minipage}[c]{.53\linewidth}
            \includegraphics[width=\linewidth]{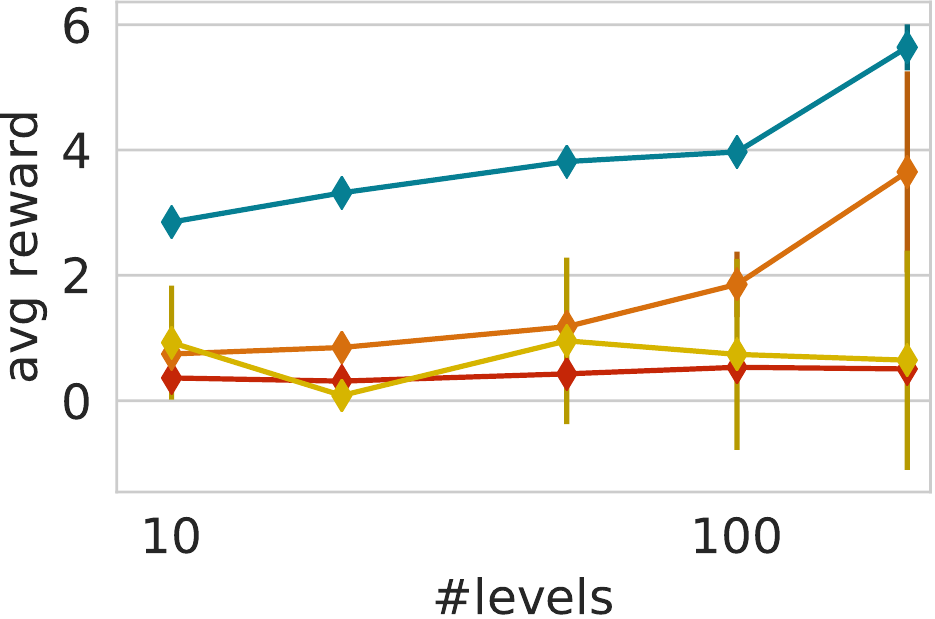}
    \end{minipage}\quad
    \begin{minipage}[c]{.18\linewidth}
            \includegraphics[width=\linewidth]{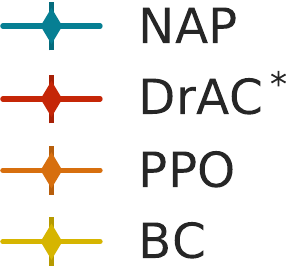}
    \end{minipage}\vspace{.3em}

    \begin{tabular}{@{}c@{\hspace{3mm}}c@{}}
        \multicolumn{2}{c}{(b) Short-horizon plans} \\
        \includegraphics[width=0.35\linewidth]{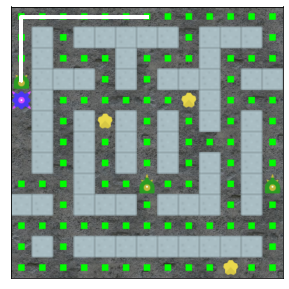}&
        \includegraphics[width=0.35\linewidth]{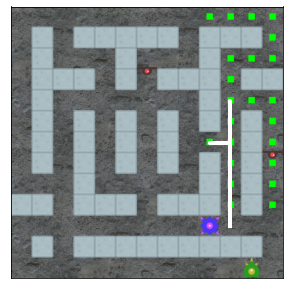}\\
    \end{tabular}\vspace*{-.5em}

    \caption{Performance of \method{} against PPO on the \env{Chaser} environment (a) trained on 10, 20, 50, 100 and 200 levels. In (b) we show the short-horizon plans (white) of the agent (blue) at step 5 and 110 in the environment.}
    \label{fig:chaser-performance}
\end{figure}
\begin{figure*}
    \newlength{\hpich}
    \setlength{\hpich}{0.164\linewidth}
    \centering
    \begin{tabular}{@{}c@{\ \ }c@{\ \ }c@{\ \ }c@{\ \ }}
         \textsc{Crash 5x5} &          \textsc{Crash 5x10}  & \textsc{Maze} & \textsc{Leaper(Grid)} \\
         \includegraphics[height=\hpich]{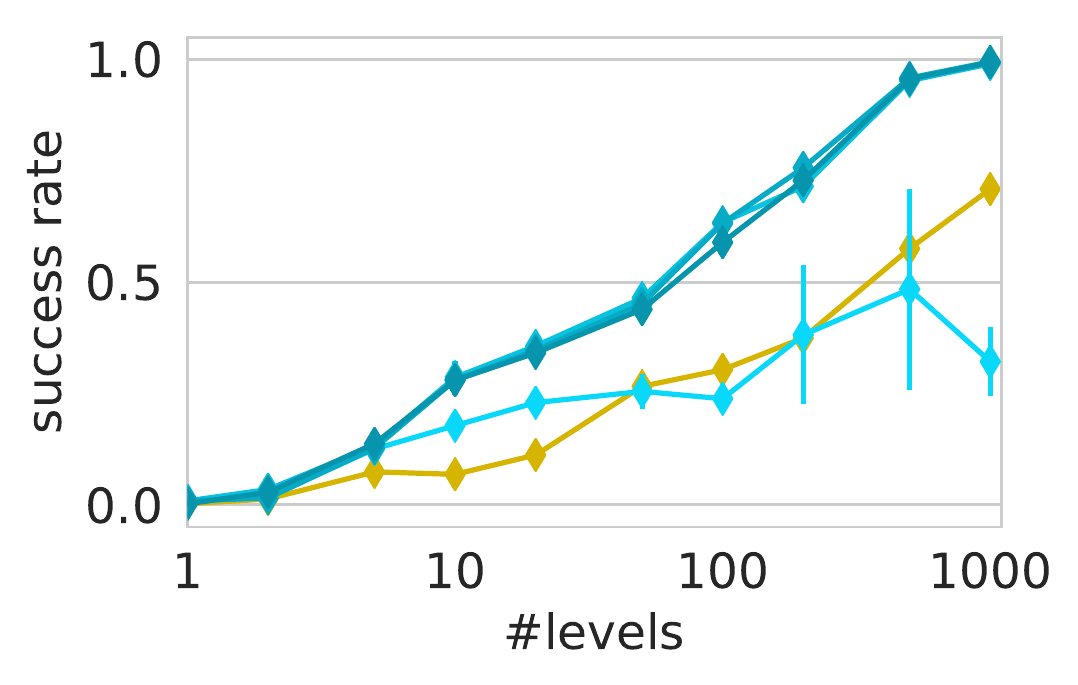}
         &
         \includegraphics[height=\hpich]{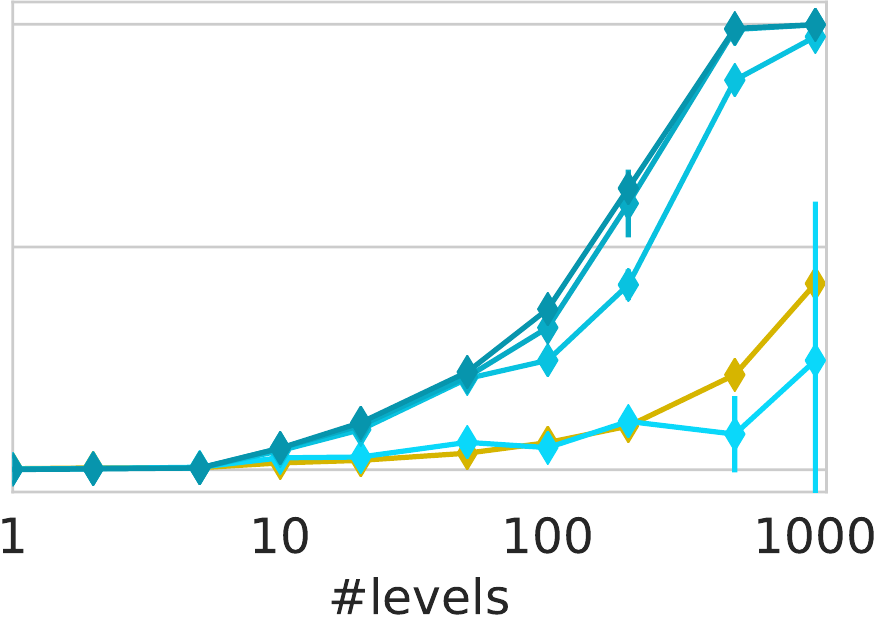}
         &
         \includegraphics[height=\hpich]{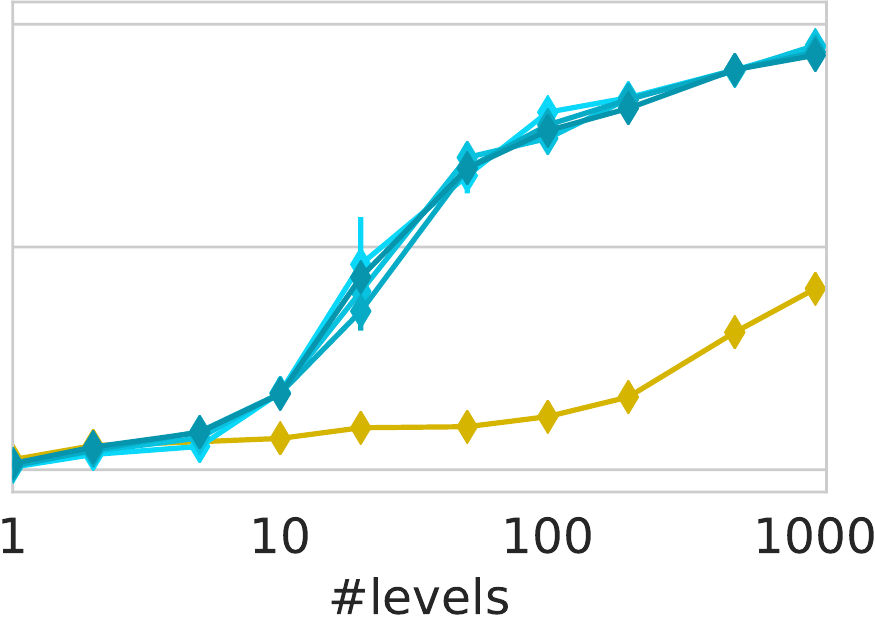}
         &
         \includegraphics[height=\hpich]{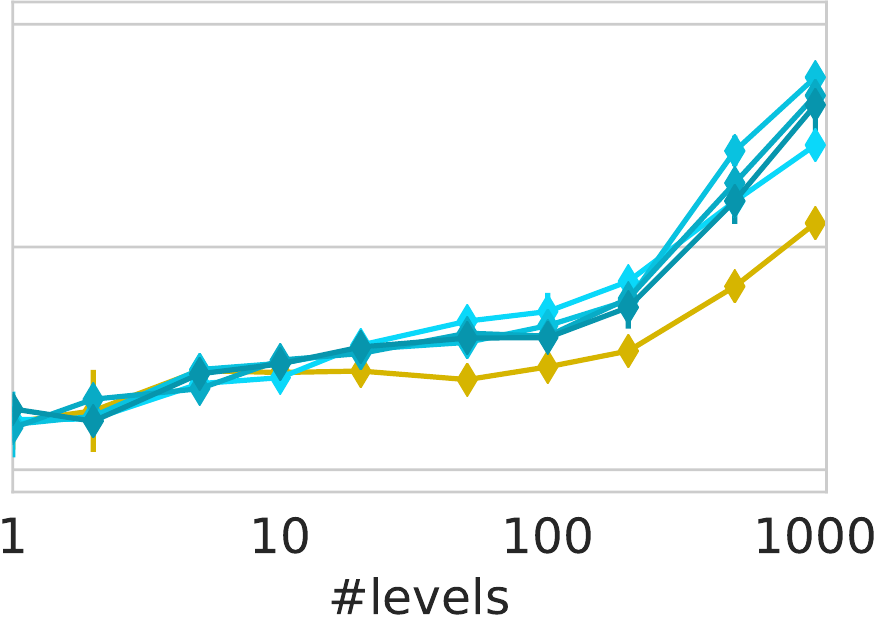}\\
         \multicolumn{4}{c}{
             \includegraphics[width=.5\linewidth]{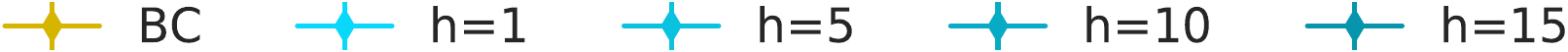}
         }
    \end{tabular}\vspace*{-.8em}
    \caption{Test success rate of our method with different horizon lengths. The solver assumes that the last horizon step costs remain to infinity. In this sense, the horizon of length 1 corresponds to a static solver.}
    \label{fig:horizon-plots}
\end{figure*}
In \fig{fig:performance-plots}, we show the performance of the methods when exposed to a different number of levels at training time.
As reported in \cite{CobbeEtAL2019:procgen},
 the baselines have a large generalization gap, and also poor performance when $<10\,000$ levels are seen.
We find that \method{} shows strong generalization performance already for $<500$ levels.
In some environments, such as \env{Maze} we obtain near $80\%$ success rate already with just $100$ levels which is never reached by PPO or \drac{} even after seeing a $1000$ levels.
Our \method{} trained on $1000$ levels reaches the same performance as PPO trained on $200\,000$ levels of the \env{Maze} environment.
For \Crash{} $5\times 5$, already with 30 trajectories a third of the $1000$ test-levels can be solved, while the behavior cloning baseline manages less than $50$ out of the $1000$.




\subsection{Sensitivity to the Planning Horizon}

We provide a sensitivity analysis of the performance with different planning horizons.
Our results indicate that longer horizons benefit environments with increased dynamical interactions.
As apparent from \fig{fig:horizon-plots}, our method outperforms the behavior cloning baseline in all of the environments, the gap between the methods being correlated with the complexity of the environment.
It can also be seen that making the planning horizon smaller in environments with dynamics hurts performance.

On the other hand, for environments with no dynamics, such as the maze environment, there is no benefit in using a longer planning horizon, as expected.
Nevertheless, there is still strong performance gain in generalization when using \method{} as opposed to vanilla imitation learning from expert trajectories thanks to the introduced combinatorial inductive bias.

\subsection{Path Optimality}
\begin{figure}[h]
    \newlength{\boxpic}
    \setlength{\boxpic}{0.4\linewidth}
   \centering
    \begin{tabular}{c@{}c@{}}
         \hspace{0.13\linewidth}\textsc{Leaper(Grid)} & \textsc{Maze} \\
         \includegraphics[height=\boxpic]{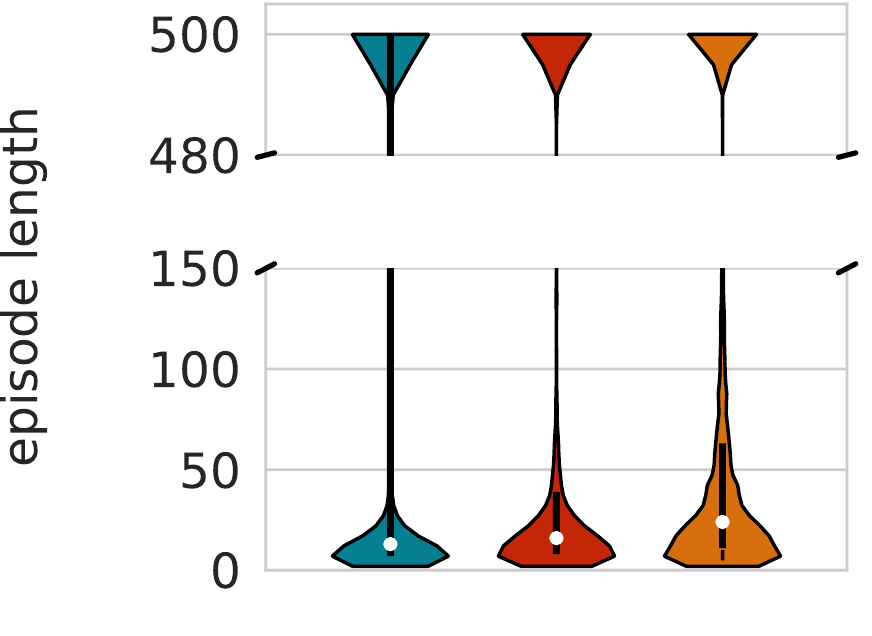}
         &
         \includegraphics[height=\boxpic]{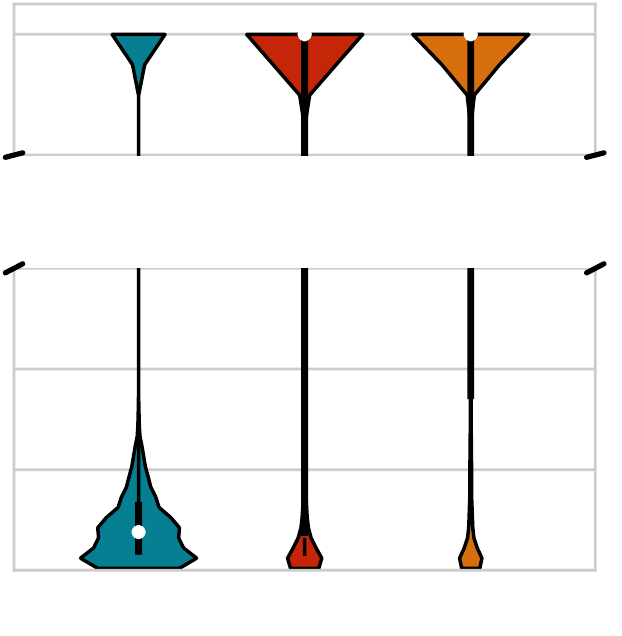}\\
         \multicolumn{2}{c}{
             \hspace{0.13\linewidth}\includegraphics[width=.6\linewidth]{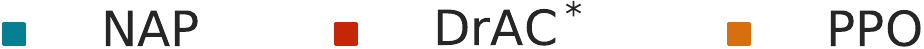}
         }
    \end{tabular}
    \caption{Distributions of test level episode lengths after training on $500$ levels for the \textsc{Leaper(Grid)} and \textsc{Maze} environments over 3 seeds. We observe that \method{} tends to take shorter paths than \drac$^*$ and PPO. We also observe that the median episode length for the baselines in the \env{Maze} task is located in the upper part, which corresponds to unsolved levels. }
    \label{fig:violin-plots}
\end{figure}

\goodbreak
We have observed that introducing combinatorial structural biases with a method such as \method{} can benefit generalization greatly.
Another interesting question is how optimal the paths taken by \method{} are.
Even though \method{} has an advantage in having access to expert optimal or near-optimal trajectories, this doesn't necessarily translate to being able to act optimally in unseen situations.

For this analysis we have plotted the episode length distribution across runs for \method{}, \drac$^*$ and PPO, which can be seen in  \fig{fig:violin-plots}.
This shows us that \method{} generalizes with more optimal (shorter) paths in comparison to \drac$^*$  and PPO, even when compared to levels solved by all of the methods.
Even in the \textsc{Leaper(Grid)} environment, where \drac$^*$  outperforms \method{} by measure of success rate, \method{} still outperforms \drac$^*$  in terms of the path lengths.

\section{Discussion}
We have shown that hybrid neuro-algorithmic policies consisting of deep feature extraction and a shortest path solver -- made differentiable via blackbox differentiation~\citep{VlastelicaEtal2020} -- \emph{enable learning policies that generalize to unseen environment settings in the low-data regime}.
Hybrid architectures are a stepping stone towards use of better inductive biases that enable stronger generalization.
In \method, the inductive bias that we impose is the topology of the latent planning graph in conjunction with a planning algorithm.
Introducing the shortest-path solver as a module shifts the combinatorial complexity of the planning problem to efficient algorithmic implementations while alleviating the learning of good representations for planning.

Although there is a clear benefit in using \method, the method comes with certain caveats.
We assume that the topological structure (\ie that there is an underlying grid structure with a set of 5 actions) of the latent planning graph is known a priori.
Furthermore, we assume that the structure of the latent graph is fixed and not dynamically changing over time, \ie that each available action at a vertex corresponds to the same edge.
Any results allowing to abandon some of these assumption will vastly increase applicability of this method and should be of immediate interest.


\section{Acknowledgment}
We thank the International Max Planck Research School for Intelligent Systems (IMPRS-IS) for supporting Marin Vlastelica. We acknowledge the support from the German Federal Ministry of Education and Research (BMBF) through the Tübingen AI Center (FKZ: 01IS18039B).

\bibliography{main}

\begin{thebibliography}{}

\bibitem[Abbeel and Ng, 2004]{abbeel2004apprenticeship}
Abbeel, P. and Ng, A.~Y. (2004).
\newblock Apprenticeship learning via inverse reinforcement learning.
\newblock In {\em Proceedings of the twenty-first international conference on
  Machine learning}, page~1.

\bibitem[Aghasadeghi and Bretl, 2011]{aghasadeghi2011maximum}
Aghasadeghi, N. and Bretl, T. (2011).
\newblock Maximum entropy inverse reinforcement learning in continuous state
  spaces with path integrals.
\newblock In {\em 2011 IEEE/RSJ International Conference on Intelligent Robots
  and Systems}, pages 1561--1566. IEEE.

\bibitem[Amos et~al., 2017]{amos2017input}
Amos, B., Xu, L., and Kolter, J.~Z. (2017).
\newblock Input convex neural networks.
\newblock In {\em International Conference on Machine Learning}, pages
  146--155.

\bibitem[Amos and Yarats, 2020]{br2019differentiable}
Amos, B. and Yarats, D. (2020).
\newblock The differentiable cross-entropy method.
\newblock In {\em International Conference on Machine Learning, ICML}.

\bibitem[anu et~al., 2020]{raileanu2020automatic}
anu, R., Goldstein, M., Yarats, D., Kostrikov, I., and Fergus, R. (2020).
\newblock Automatic data augmentation for generalization in deep reinforcement
  learning.
\newblock {\em arXiv preprint arXiv:2006.12862}.

\bibitem[Asai and Fukunaga, 2017]{asai2017classical}
Asai, M. and Fukunaga, A. (2017).
\newblock Classical planning in deep latent space: Bridging the
  subsymbolic-symbolic boundary.
\newblock {\em arXiv preprint arXiv:1705.00154}.

\bibitem[Asai and Kajino, 2019]{asai2019towards}
Asai, M. and Kajino, H. (2019).
\newblock Towards stable symbol grounding with zero-suppressed state
  autoencoder.
\newblock In {\em Proceedings of the International Conference on Automated
  Planning and Scheduling}, volume~29, pages 592--600.

\bibitem[Berthet et~al., 2020]{berthet2020learning}
Berthet, Q., Blondel, M., Teboul, O., Cuturi, M., Vert, J.-P., and Bach, F.
  (2020).
\newblock Learning with differentiable perturbed optimizers.
\newblock {\em arXiv preprint arXiv:2002.08676}.

\bibitem[Bharadhwaj et~al., 2020]{bharadhwaj2020model}
Bharadhwaj, H., Xie, K., and Shkurti, F. (2020).
\newblock Model-predictive control via cross-entropy and gradient-based
  optimization.
\newblock {\em arXiv preprint arXiv:2004.08763}.

\bibitem[Blaes et~al., 2019]{blaes2019control}
Blaes, S., Vlastelica~Pogan\v{c}i\'{c}, M., Zhu, J., and Martius, G. (2019).
\newblock Control what you can: Intrinsically motivated task-planning agent.
\newblock In {\em Advances in Neural Information Processing Systems 32}, pages
  12541--12552. Curran Associates, Inc.

\bibitem[Blundell et~al., 2016]{blundell2016model}
Blundell, C., Uria, B., Pritzel, A., Li, Y., Ruderman, A., Leibo, J.~Z., Rae,
  J., Wierstra, D., and Hassabis, D. (2016).
\newblock Model-free episodic control.
\newblock {\em arXiv preprint arXiv:1606.04460}.

\bibitem[Burget et~al., 2016]{burget2016bi}
Burget, F., Bennewitz, M., and Burgard, W. (2016).
\newblock Bi 2 rrt*: An efficient sampling-based path planning framework for
  task-constrained mobile manipulation.
\newblock In {\em 2016 IEEE/RSJ International Conference on Intelligent Robots
  and Systems (IROS)}, pages 3714--3721. IEEE.

\bibitem[Chen et~al., 2018]{chen2018optimal}
Chen, Y., Shi, Y., and Zhang, B. (2018).
\newblock Optimal control via neural networks: A convex approach.
\newblock In {\em International Conference on Learning Representations}.

\bibitem[Cobbe et~al., 2019]{CobbeEtAL2019:procgen}
Cobbe, K., Hesse, C., Hilton, J., and Schulman, J. (2019).
\newblock Leveraging procedural generation to benchmark reinforcement learning.
\newblock {\em arXiv preprint:1912.01588}.

\bibitem[Daw et~al., 2005]{daw2005uncertainty}
Daw, N.~D., Niv, Y., and Dayan, P. (2005).
\newblock Uncertainty-based competition between prefrontal and dorsolateral
  striatal systems for behavioral control.
\newblock {\em Nature neuroscience}, 8(12):1704--1711.

\bibitem[Demirovic et~al., 2019]{predict-optimize-ranking}
Demirovic, E., Stuckey, P.~J., Bailey, J., Chan, J., Leckie, C., Ramamohanarao,
  K., and Guns, T. (2019).
\newblock Predict+optimise with ranking objectives: Exhaustively learning
  linear functions.
\newblock In Kraus, S., editor, {\em Proceedings of the Twenty-Eighth
  International Joint Conference on Artificial Intelligence, {IJCAI} 2019,
  Macao, China, August 10-16, 2019}, pages 1078--1085. ijcai.org.

\bibitem[Dijkstra, 1959]{Dijkstra}
Dijkstra, E.~W. (1959).
\newblock A note on two problems in connexion with graphs.
\newblock {\em Numer. Math.}, 1(1):269--271.

\bibitem[Djolonga and Krause, 2017]{djolonga2017differentiable}
Djolonga, J. and Krause, A. (2017).
\newblock Differentiable learning of submodular models.
\newblock In {\em Advances in Neural Information Processing Systems}, pages
  1013--1023.

\bibitem[Elmachtoub and Grigas, 2017]{predict-optimize}
Elmachtoub, A.~N. and Grigas, P. (2017).
\newblock Smart "predict, then optimize".
\newblock {\em ArXiv}, abs/1710.08005.

\bibitem[Eysenbach et~al., 2019]{eysenbach2019search}
Eysenbach, B., Salakhutdinov, R.~R., and Levine, S. (2019).
\newblock Search on the replay buffer: Bridging planning and reinforcement
  learning.
\newblock In {\em Advances in Neural Information Processing Systems 32}, pages
  15246--15257. Curran Associates, Inc.

\bibitem[Ferber et~al., 2020]{ferber2020mipaal}
Ferber, A., Wilder, B., Dilkina, B., and Tambe, M. (2020).
\newblock Mipaal: Mixed integer program as a layer.
\newblock In {\em AAAI}, pages 1504--1511.

\bibitem[Finn and Levine, 2017]{finn2017deep}
Finn, C. and Levine, S. (2017).
\newblock Deep visual foresight for planning robot motion.
\newblock In {\em 2017 IEEE International Conference on Robotics and Automation
  (ICRA)}, pages 2786--2793. IEEE.

\bibitem[Gammell et~al., 2014]{gammell2014informed}
Gammell, J.~D., Srinivasa, S.~S., and Barfoot, T.~D. (2014).
\newblock Informed rrt*: Optimal sampling-based path planning focused via
  direct sampling of an admissible ellipsoidal heuristic.
\newblock In {\em 2014 IEEE/RSJ International Conference on Intelligent Robots
  and Systems}, pages 2997--3004. IEEE.

\bibitem[Hafner et~al., 2019]{Hafner2019:PlaNet}
Hafner, D., Lillicrap, T., Fischer, I., Villegas, R., Ha, D., Lee, H., and
  Davidson, J. (2019).
\newblock Learning latent dynamics for planning from pixels.
\newblock In {\em International Conference on Machine Learning}, volume~97 of
  {\em ICML'19}, pages 2555--2565, Long Beach, California, USA.

\bibitem[Ho and Ermon, 2016]{ho2016generative}
Ho, J. and Ermon, S. (2016).
\newblock Generative adversarial imitation learning.
\newblock In {\em Advances in neural information processing systems}, pages
  4565--4573.

\bibitem[Janner et~al., 2019]{janner2019trust}
Janner, M., Fu, J., Zhang, M., and Levine, S. (2019).
\newblock When to trust your model: Model-based policy optimization.
\newblock In {\em Advances in Neural Information Processing Systems}, pages
  12519--12530.

\bibitem[Kaiser et~al., 2019]{kaiser2019model}
Kaiser, L., Babaeizadeh, M., Milos, P., Osinski, B., Campbell, R.~H.,
  Czechowski, K., Erhan, D., Finn, C., Kozakowski, P., Levine, S., et~al.
  (2019).
\newblock Model-based reinforcement learning for atari.
\newblock {\em arXiv preprint arXiv:1903.00374}.

\bibitem[Karkus et~al., 2019]{karkus2019differentiable}
Karkus, P., Ma, X., Hsu, D., Kaelbling, L.~P., Lee, W.~S., and
  Lozano{-}P{\'{e}}rez, T. (2019).
\newblock Differentiable algorithm networks for composable robot learning.
\newblock In Bicchi, A., Kress{-}Gazit, H., and Hutchinson, S., editors, {\em
  Robotics: Science and Systems XV, University of Freiburg, Freiburg im
  Breisgau, Germany, June 22-26, 2019}.

\bibitem[Kumar et~al., 2019]{kumar2019LEGOLE}
Kumar, R., Mandalika, A., Choudhury, S., and Srinivasa, S. (2019).
\newblock Lego: Leveraging experience in roadmap generation for sampling-based
  planning.
\newblock {\em 2019 IEEE/RSJ International Conference on Intelligent Robots and
  Systems (IROS)}, pages 1488--1495.

\bibitem[Kuo et~al., 2018]{kuo2018deep}
Kuo, Y.-L., Barbu, A., and Katz, B. (2018).
\newblock Deep sequential models for sampling-based planning.
\newblock In {\em 2018 IEEE/RSJ International Conference on Intelligent Robots
  and Systems (IROS)}, pages 6490--6497. IEEE.

\bibitem[Li et~al., 2020]{LiTorralba2020:CompKoopman}
Li, Y., He, H., Wu, J., Katabi, D., and Torralba, A. (2020).
\newblock Learning compositional koopman operators for model-based control.
\newblock In {\em International Conference on Learning Representations}.

\bibitem[Mandi et~al., 2019]{predict-and-optimize-comb}
Mandi, J., Demirovic, E., Stuckey, P.~J., and Guns, T. (2019).
\newblock Smart predict-and-optimize for hard combinatorial optimization
  problems.
\newblock {\em CoRR}, abs/1911.10092.

\bibitem[Nachum et~al., 2018]{nachum2018data}
Nachum, O., Gu, S.~S., Lee, H., and Levine, S. (2018).
\newblock Data-efficient hierarchical reinforcement learning.
\newblock In {\em Advances in Neural Information Processing Systems}, pages
  3303--3313.

\bibitem[Neu and Szepesv{\'a}ri, 2012]{neu2012apprenticeship}
Neu, G. and Szepesv{\'a}ri, C. (2012).
\newblock Apprenticeship learning using inverse reinforcement learning and
  gradient methods.
\newblock {\em arXiv preprint arXiv:1206.5264}.

\bibitem[Ng and Russell, 2000]{ng2000algorithms}
Ng, A.~Y. and Russell, S. (2000).
\newblock Algorithms for inverse reinforcement learning.
\newblock In {\em in Proc. 17th International Conf. on Machine Learning}.
  Citeseer.

\bibitem[Niculae et~al., 2018]{niculae2018sparsemap}
Niculae, V., Martins, A.~F., Blondel, M., and Cardie, C. (2018).
\newblock Sparsemap: Differentiable sparse structured inference.
\newblock {\em arXiv preprint arXiv:1802.04223}.

\bibitem[Oh et~al., 2017]{oh2017value}
Oh, J., Singh, S., and Lee, H. (2017).
\newblock Value prediction network.
\newblock In {\em Advances in Neural Information Processing Systems}, pages
  6118--6128.

\bibitem[Pinneri et~al., 2021]{pinneri2021:strong-policies}
Pinneri, C., Sawant, S., Blaes, S., and Martius, G. (2021).
\newblock Extracting strong policies for robotics tasks from zero-order
  trajectory optimizers.
\newblock In {\em International Conference on Learning Representations}, ICLR.

\bibitem[Racani{\`e}re et~al., 2017]{racaniere2017imagination}
Racani{\`e}re, S., Weber, T., Reichert, D., Buesing, L., Guez, A.,
  Jimenez~Rezende, D., Puigdom{\`e}nech~Badia, A., Vinyals, O., Heess, N., Li,
  Y., et~al. (2017).
\newblock Imagination-augmented agents for deep reinforcement learning.
\newblock {\em Advances in neural information processing systems},
  30:5690--5701.

\bibitem[Reddy et~al., 2019]{reddy2019sqil}
Reddy, S., Dragan, A.~D., and Levine, S. (2019).
\newblock Sqil: Imitation learning via reinforcement learning with sparse
  rewards.
\newblock {\em arXiv preprint arXiv:1905.11108}.

\bibitem[Rol\'inek et~al., 2020]{rolinek2020cvpr}
Rol\'inek, M., Musil, V., Paulus, A., Vlastelica, M., Michaelis, C., and
  Martius, G. (2020).
\newblock Optimizing ranking-based metrics with blackbox differentiation.
\newblock In {\em Conference on Computer Vision and Pattern Recognition},
  CVPR'20.

\bibitem[Ross et~al., 2011]{ross2011reduction}
Ross, S., Gordon, G., and Bagnell, D. (2011).
\newblock A reduction of imitation learning and structured prediction to
  no-regret online learning.
\newblock In {\em Proceedings of the fourteenth international conference on
  artificial intelligence and statistics}, pages 627--635.

\bibitem[Savinov et~al., 2018]{savinov2018semi}
Savinov, N., Dosovitskiy, A., and Koltun, V. (2018).
\newblock Semi-parametric topological memory for navigation.
\newblock In {\em International Conference on Learning Representations}.

\bibitem[Schaul et~al., 2015]{schaul2015:UVFA}
Schaul, T., Horgan, D., Gregor, K., and Silver, D. (2015).
\newblock Universal value function approximators.
\newblock In {\em International Conference on Machine Learning}, volume~37 of
  {\em ICML}, pages 1312--1320.

\bibitem[Schrittwieser et~al., 2019]{schrittwieser2019mastering}
Schrittwieser, J., Antonoglou, I., Hubert, T., Simonyan, K., Sifre, L.,
  Schmitt, S., Guez, A., Lockhart, E., Hassabis, D., Graepel, T., et~al.
  (2019).
\newblock Mastering atari, go, chess and shogi by planning with a learned
  model.
\newblock {\em arXiv preprint arXiv:1911.08265}.

\bibitem[Silver et~al., 2017a]{silver2017predictron}
Silver, D., Hasselt, H., Hessel, M., Schaul, T., Guez, A., Harley, T.,
  Dulac-Arnold, G., Reichert, D., Rabinowitz, N., Barreto, A., et~al. (2017a).
\newblock The predictron: End-to-end learning and planning.
\newblock In {\em International Conference on Machine Learning}, pages
  3191--3199. PMLR.

\bibitem[Silver et~al., 2017b]{Silver2017:end2endlearningandplanning}
Silver, D., van Hasselt, H., Hessel, M., Schaul, T., Guez, A., Harley, T.,
  Dulac-Arnold, G., Reichert, D., Rabinowitz, N., Barreto, A., and Degris, T.
  (2017b).
\newblock The predictron: End-to-end learning and planning.
\newblock In {\em International Conference on Machine Learning}, volume~70 of
  {\em ICML'17}, pages 3191--3199. PMLR.

\bibitem[Srinivas et~al., 2018]{Srinivas2018:UPN}
Srinivas, A., Jabri, A., Abbeel, P., Levine, S., and Finn, C. (2018).
\newblock Universal planning networks: Learning generalizable representations
  for visuomotor control.
\newblock In {\em International Conference on Machine Learning}, volume~80 of
  {\em ICML'18}, pages 4732--4741, Stockholmsmässan, Stockholm Sweden.

\bibitem[Sutton, 1991]{sutton1991dyna}
Sutton, R.~S. (1991).
\newblock Dyna, an integrated architecture for learning, planning, and
  reacting.
\newblock {\em ACM Sigart Bulletin}, 2(4):160--163.

\bibitem[Sutton et~al., 2008]{sutton2008dyna}
Sutton, R.~S., Szepesv{\'a}ri, C., Geramifard, A., and Bowling, M. (2008).
\newblock Dyna-style planning with linear function approximation and
  prioritized sweeping.
\newblock In {\em Proceedings of the Twenty-Fourth Conference on Uncertainty in
  Artificial Intelligence}, pages 528--536.

\bibitem[Tamar et~al., 2016]{tamar2016value}
Tamar, A., Wu, Y., Thomas, G., Levine, S., and Abbeel, P. (2016).
\newblock Value iteration networks.
\newblock In {\em Advances in Neural Information Processing Systems}, pages
  2154--2162.

\bibitem[Vlastelica et~al., 2020]{VlastelicaEtal2020}
Vlastelica, M., Paulus, A., Musil, V., Martius, G., and Rol{\'{\i}}nek, M.
  (2020).
\newblock Differentiation of blackbox combinatorial solvers.
\newblock In {\em International Conference on Learning Representations},
  ICLR'20.

\bibitem[Wang et~al., 2019]{wang2019satnet}
Wang, P.-W., Donti, P.~L., Wilder, B., and Kolter, Z. (2019).
\newblock Satnet: Bridging deep learning and logical reasoning using a
  differentiable satisfiability solver.
\newblock {\em arXiv preprint arXiv:1905.12149}.

\bibitem[Wu et~al., 2020]{wu2020scalable}
Wu, G., Say, B., and Sanner, S. (2020).
\newblock Scalable planning with deep neural network learned transition models.
\newblock {\em Journal of Artificial Intelligence Research}, 68:571--606.

\bibitem[Yao et~al., 2009]{yao2009multi}
Yao, H., Bhatnagar, S., Diao, D., Sutton, R.~S., and Szepesv{\'a}ri, C. (2009).
\newblock Multi-step dyna planning for policy evaluation and control.
\newblock In {\em Advances in neural information processing systems}, pages
  2187--2195.

\bibitem[Yonetani et~al., 2020]{yonetani2020path}
Yonetani, R., Taniai, T., Barekatain, M., Nishimura, M., and Kanezaki, A.
  (2020).
\newblock Path planning using neural a* search.
\newblock {\em arXiv preprint arXiv:2009.07476}.

\end{thebibliography}


\begin{thebibliography}{}

\bibitem[Kostrikov et~al., 2020]{kostrikov2020image}
Kostrikov, I., Yarats, D., and Fergus, R. (2020).
\newblock Image augmentation is all you need: Regularizing deep reinforcement
  learning from pixels.
\newblock {\em arXiv preprint arXiv:2004.13649}.

\bibitem[Laskin et~al., 2020]{laskin2020reinforcement}
Laskin, M., Lee, K., Stooke, A., Pinto, L., Abbeel, P., and Srinivas, A.
  (2020).
\newblock Reinforcement learning with augmented data.
\newblock {\em arXiv preprint arXiv:2004.14990}.

\bibitem[Srinivas et~al., 2020]{srinivas2020curl}
Srinivas, A., Laskin, M., and Abbeel, P. (2020).
\newblock Curl: Contrastive unsupervised representations for reinforcement
  learning.
\newblock {\em arXiv preprint arXiv:2004.04136}.

\bibitem[Tobin et~al., 2017]{tobin2017domain}
Tobin, J., Fong, R., Ray, A., Schneider, J., Zaremba, W., and Abbeel, P.
  (2017).
\newblock Domain randomization for transferring deep neural networks from
  simulation to the real world.
\newblock In {\em 2017 IEEE/RSJ international conference on intelligent robots
  and systems (IROS)}, pages 23--30. IEEE.

\bibitem[Zhan et~al., 2020]{zhan2020framework}
Zhan, A., Zhao, P., Pinto, L., Abbeel, P., and Laskin, M. (2020).
\newblock A framework for efficient robotic manipulation.
\newblock {\em arXiv preprint arXiv:2012.07975}.

\end{thebibliography}
\bibliographystyle{apalike}


\newpage
\clearpage

\appendix

\pagestyle{empty}
\twocolumn[
\begin{center}
{\titlefont \textbf{Supplemetary Material for \papertitle}}
\end{center}
\vspace{3em}
]

\section{Data Generation}
\label{sec:app-data-gen}

To do imitation learning, we require expert data.
For \env{Crash Jewel Hunt} (\env{Crash $5\times 5$} and \env{$5\times 10$}), \env{Leaper(Grid)} and \env{Maze} we can determine the exact ground truth costs leading to optimal behavior.
As an example, \env{Crash $5\times 5$} contains moving boxes that when encountered lead to instant death, meaning infinite costs and otherwise the fixed cost of moving around in the environment.

Since the environments become deterministic for a fixed random seed, we first unrolled their dynamics for each level.
After obtaining the underlying grid structure and entities, we labeled them with costs and constructed a graph that reflects the grid structure.
An expert trajectory is constructed by applying Dijkstra's algorithm on this graph and the human-labeled costs and then executing in simulation.

For the \env{Crash Jewel Hunt} experiments, we randomly sampled $2000$ solvable levels by varying number of boxes per column, their speed, the agent start position and the jewel position. The training levels were taken from the first half and the second half of levels was used for testing.
For the \textsc{ProcGen} environments \env{Leaper(Grid)} and \env{Maze} we have taken the levels determined by seeds 0-1000. 

For \env{Chaser}, we applied a similar procedure but additionally, we recorded two sets of human trajectories, as we observed benefits in performance by incorporating more different expert trajectories for the same level. Since both the search procedure and human labeling are time consuming for this environment, we collected fewer expert trajectories for the \env{Chaser} than for the other environments,  $3\times100$, two-thirds of which are from human players.


Level seeds 1000000-1001000 were taken for testing in the \textsc{ProcGen} experiments.

\section{Environments}

\begin{table*}[t!]
\begin{center}

    \begin{tabular}{c|c c c c}
        &  \env{Crash $5\times 5$} & \env{Crash $5\times 10$} & \env{Leaper(Grid)} & \env{Maze} \\
        \hline \\
        learning rate &  $10^{-3}$ & $10^{-3}$ & $10^{-3}$ & $10^{-3}$ \\
        $\alpha$ &  0.2 & 0.2 & 0.15 & 0.15 \\
        $\lambda$ & 20 & 20 & 20 & 20 \\
        resnet layers & 4 & 4 & 4 & 4 \\
        kernel size & 4 & 4 & 6 & 6 \\
        batch size & 32 & 32 & 16 & 16 \\
    \end{tabular}
    \caption{Training hyperparameters, where $alpha$ denotes the margin that was used on the vertex costs and $\lambda$ the interpolation parameter for blackbox differentiation of Dijkstra's algorithm. We vary the kernel size of the initial convolution for ResNet18.}
    \label{tab:hyperparams-1}
\end{center}
\end{table*}

Our method is applicable in discrete environments, therefore we evaluated on environments from the \env{ProcGen} benchmark and the \env{Crash Jewel Hunt} environment.

We created the \env{Crash Jewel Hunt} environment to evaluate our method, where the goal is for the fox (Crash) to reach the jewel.
We found this environment convenient since we can influence the combinatorial difficulty directly, which is not true for the \env{ProcGen} benchmark where we are limited to the random seeds used in the OpenAI implementation.
The sources of variation in the \env{Crash Jewel Hunt} are the box velocities, initial positions, sizes, as well as the agent initial position and the jewel position.

We modified the \env{Leaper} environment to make grid steps for our method to be applicable. This involved making the logs on the river move in discrete steps as well as the agent. 
Moreover, in our version, the agent is not transported by the logs as they move, but has to move actively with them. 
For an additional description of the \env{ProcGen} environmnets, we refer the reader to \citet{CobbeEtAL2019:procgen}.

\section{Network Architecture and Input}
\label{app:architecture}
For all of our experiments, we use the PyTorch implementation of the ResNet18 architecture as the base of our model.
All approaches receive two stacked frames of the two previous time steps as input to make dynamics prediction possible.
For the PPO baseline, we did not observe any benefit in adding the stacked frames as input and we used stable-baselines implementation from OpenAI to train it on the \env{ProcGen} environments.

In the case of the behavior cloning baseline, the problem is a multi-class classification problem with the output being a multinomial distribution over actions.

For the variant \method{}$^*$, we train a cost prediction network on top of which we run Dijkstra's algorithm on the output costs of the planning graph.
This requires modifications to the original ResNet18 architecture.
We remove the linear readout of the original ResNet18 architecture and replace it with a convolutional layer of filter size 1 and adaptive max pooling layer to obtain the desired dimensions of the underlying latent planning graph.
More concretely, the output $x$ of the last ResNet18 block is followed by the following operation (as output by PyTorch) to obtain the graph costs:

{\small
    \begin{tabbing}
      Seq\=uential( \\
      \>Conv2d(256, 2, kernel\_size=(1, 1), stride=(1, 1))\\
      \> Abs() \\
      \>AdaptiveMaxPool2d(output\_size=(grid\_height, grid\_width))\\
      )
    \end{tabbing}
}

Where grid\_\{height,width\} denotes the height and width of the planning grid. For the full variant of \method{} with goal prediction and agent position prediction we have a separate position classifier that has the same base architecture as the cost prediction network with 2 additional linear readouts for the likelihoods of the latent graph vertices, more concretely (as output by PyTorch):

{ \small
\begin{tabbing}
      Seq\=uential( \\
      \>Conv2d(256, 2, kernel\_size=(1, 1), stride=(1, 1))\\
      \> Abs() \\
      \>AdaptiveMaxPool2d(output\_size=(grid\_height, grid\_width))\\
      \> Flatten() \\
      \> Linear(grid\_height $\times$ grid\_width, grid\_height $\times$ grid\_width) \\
      )
    \end{tabbing}
}

For training the position classifier, we use a standard cross-entropy loss on the likelihoods.
For \method{} with position classification, we use the ground-truth expert start and goal positions to calculate the Hamming loss of the predicted path by the solver. At evaluation time, \method{} uses the position classifier to determine the start and end vertices in the latent planning graph.

\section{Training Procedure}
\label{app:training}
\pagestyle{fancy}
\fancyhead[CO, CE]{\textbf{Supplementary for \paperrunningtitle}}

For  \env{Crash $5\times 5$},  \env{Crash $5\times 10$}, \env{Leaper(Grid)} and \env{Maze} we train the models on the same \#levels, namely 1, 2, 5, 10, 20, 50, 100, 200, 500 and 1000. We evaluate on unseen 1000 levels in order to show that \method{} exhibits superior generalization. The levels are generated as per description in section \ref{sec:app-data-gen}. Each dataset is normalize to be zero mean and unit variancesize. For each dataset size (\#levels) we run experiments with 3 random restarts (seeds for network initialization). For all experiments, we make use of the \textsc{Adam} optimizer.

We determine the number of epochs for training depending on each dataset size as $\min(150000/\text{\#levels}, 15000)$ to have roughly the same number of gradient updates in each experiment.
We take the minimum over the 2 values because for smaller number of levels a large number iterations is not necessary to achieve good performance, but for a larger number of levels it is necessary.
If we observe no error on the training set, we stop the training.

For the \env{Chaser}, the training conditions were analogous to the other environments, only of slightly smaller scale due to its higher complexity. Models were trained on 10, 20, 50, 100 and 200 levels and evaluated on 200 unseen levels.
\begin{table}[h]
\begin{center}

    \begin{tabular}{c|c}
        &  \env{Chaser} \\
        \hline \\
        learning rate &  $1e^{-3}$ \\
        $\alpha$ &  0.2 \\
        $\lambda$ & 40  \\
        resnet layers & 3  \\
        kernel size & 4\\
        batch size & 16\\
    \end{tabular}
    \caption{Training hyperparameters for the \env{Chaser} experiment, where $\alpha$ denotes the margin that was used on the vertex costs and $\lambda$ the interpolation parameter for blackbox differentiation od Dijkstra.}
    \label{tab:hyperparams-2}

\end{center}
\end{table}

\subsection{PPO Training Procedure}
The training of the PPO baseline is exactly the same as described in  \citet{CobbeEtAL2019:procgen} using the official code from \url{https://github.com/openai/train-procgen}, see \tab{tab:hyperparams-ppo} for the used parameters. The network architecture is the IMPALA-CNN.
The algorithm is trained on the specified number of levels for 200 million environments interactions gathered from 256 (instead of 64 as in \citet{CobbeEtAL2019:procgen}) to compensate for not having access to 4 parallel workers.
We report numbers for 3 independent restarts.

\begin{table}[h]
\begin{center}

    \begin{tabular}{c|c}
        learning rate &  $5e^{-4}$ \\
        $\alpha$ &  0.2 \\
        discount $\gamma$ & 0.999  \\
        entropy coefficient & 0.01  \\
        steps per update & $2^{16}$
    \end{tabular}
    \caption{PPO hyperparameters, as used in \citet{CobbeEtAL2019:procgen}.}
    \label{tab:hyperparams-ppo}

\end{center}
\end{table}

\subsection{DrAC Training Procedure}

For the DrAC algorithm, we run all versions introduced by \citet{raileanu2020automatic} (Meta-DrAC, RL2-DrAC, UCB-DrAC and DrAC-Crop) and choose the best one in the main plots, denoted as DrAC$^*$.
We used the original hyperparameters from \citet{raileanu2020automatic} and the implementation from \url{https://github.com/rraileanu/auto-drac}.
As with the other experiments, we report numbers from 3 different random seeds.

\section{On Comparing Imitation Learning to Reinforcement Learning}
\label{app:imitation-vs-rl}
We compare our method to Data Augmented Actor Critic, PPO and a behavior cloning baseline.
Arguably, since \method{} is used in an imitation learning setting, it reaps benefits from having access to expert trajectories.
Nevertheless, it is not straight forward that embedding a solver in a neural architecture leads to better generalization in comparison to reinforcement learning methods.

Behavior cloning with a standard neural architecture has access to the same amount of data, whereas reinforcement learning agents have access to orders of magnitude more data for inference ($2 \cdot 10^8$ transitions in comparison to $\sim 10^5$ max).
This would lead us to believe that reinforcement learning agents are able to generalize well because of the sheer amount of data that they have at their disposal, but we show that nevertheless it is possible to extract meaningful policies even with a small number of training levels seen with expert trajectories, that are more optimal.

\begin{figure*}[t]
    \setlength{\pich}{0.162\linewidth}
    \centering
    \begin{tabular}{c@{\ \ }c@{\ \ }c@{\ \ }c@{}}
        \textsc{Crash $5\times5$} & \textsc{Crash $5\times10$} & \textsc{Maze} & \textsc{Leaper} \\
        \includegraphics[height=\pich]{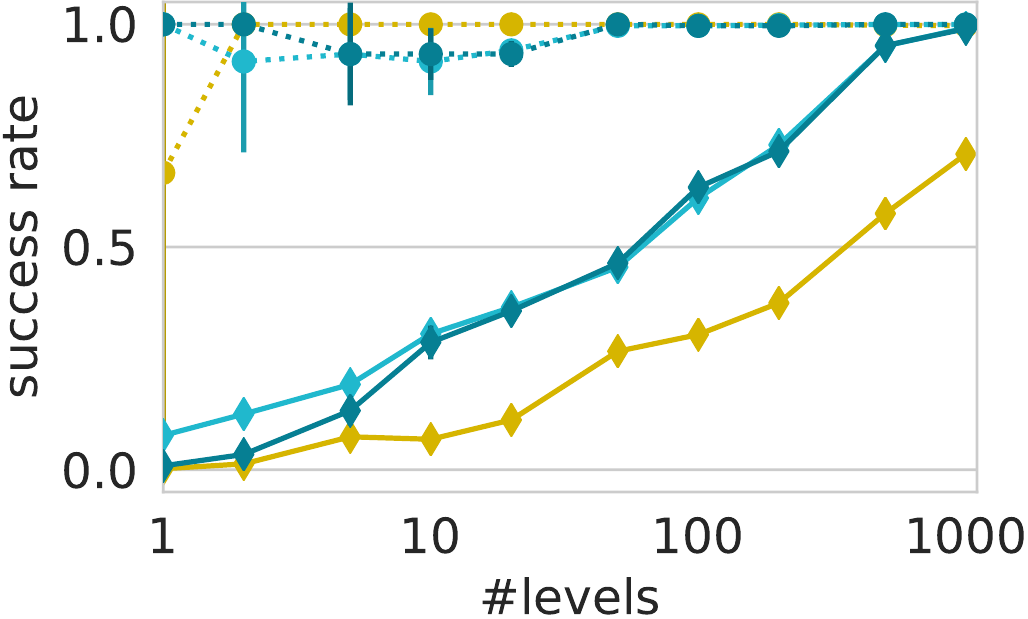} &
        \includegraphics[height=\pich]{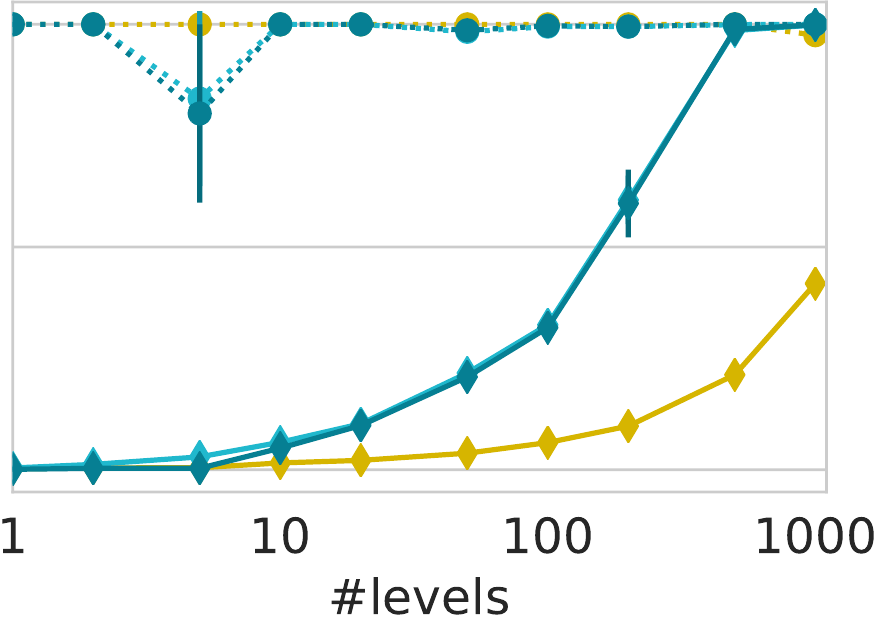} &
        \includegraphics[height=\pich]{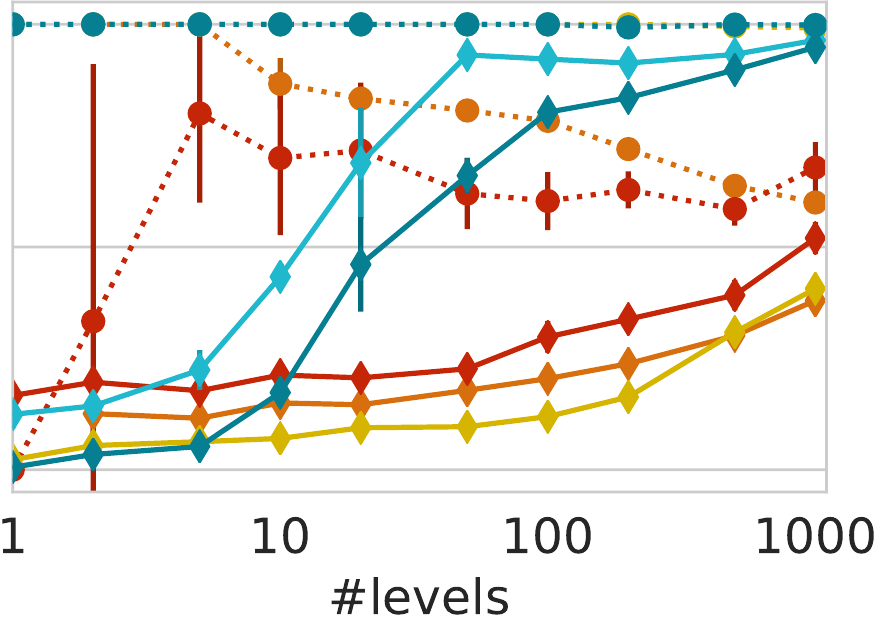} &
        \includegraphics[height=\pich]{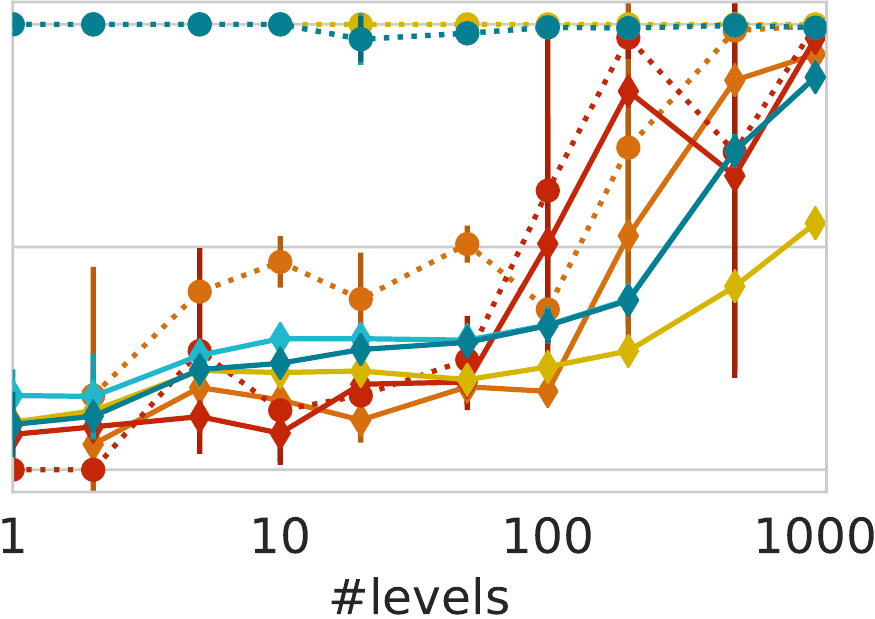} \\
        \multicolumn{4}{c}{
             \includegraphics[width=.5\linewidth]{icml/performance_plot_legend-crop.pdf}
         }
        
    \end{tabular}
    \vspace{-.7em}
    \caption{The dotted lines denote the training performance of the methods. We observe that the behavior cloning baseline and \method{} have fitted the training set almost perfectly.}
    \label{fig:supp-performance}
\end{figure*}

\begin{figure*}[t]
    \setlength{\pich}{0.12\linewidth}
    \centering
    \begin{tabularx}{\linewidth}{c@{\hspace{2em}}|Y@{\hspace{2.5em} }Y@{\ \ }Y@{\ \ }Y@{\ \ }Y@{\ \ }Y@{\ \ }X@{}}
              & \hspace{4.05em} 10  & 20 & 50 & 100 & 200 & 500 \\
         \textsc{Maze} & 
         \includegraphics[height=\pich]{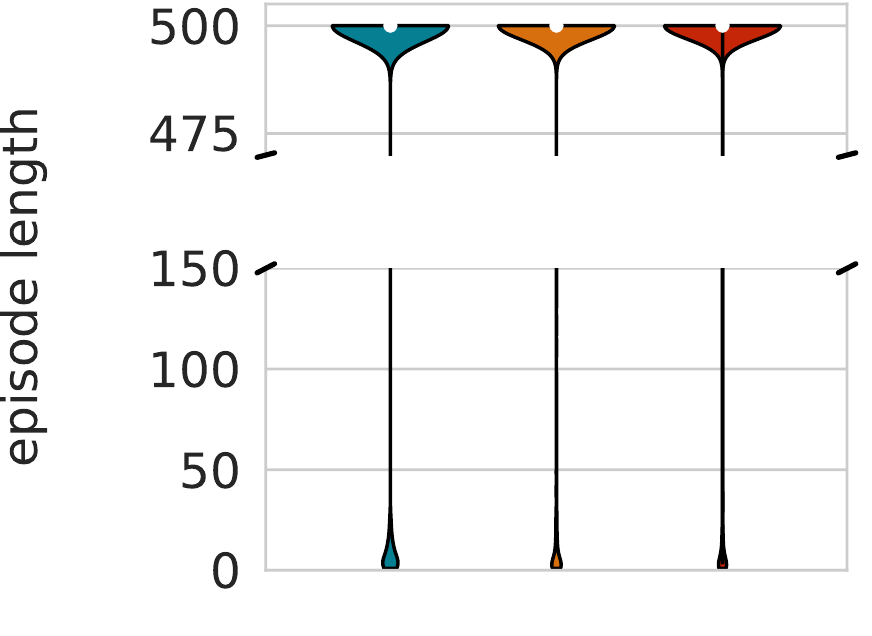} &
         \includegraphics[height=\pich]{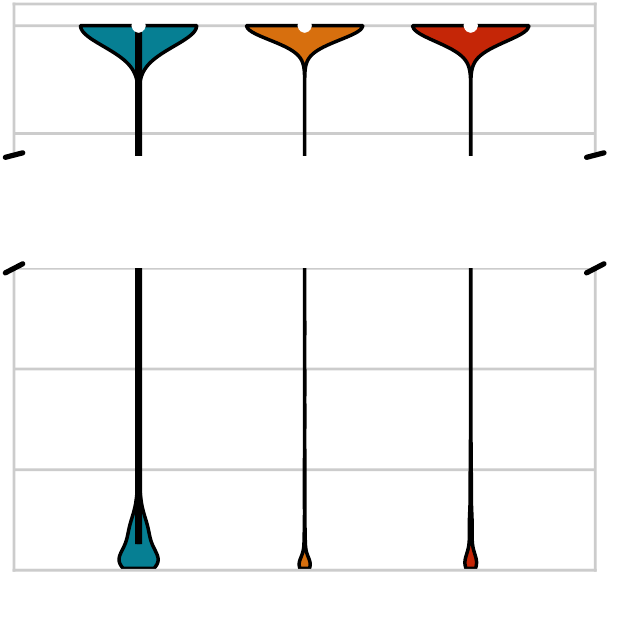} &
         \includegraphics[height=\pich]{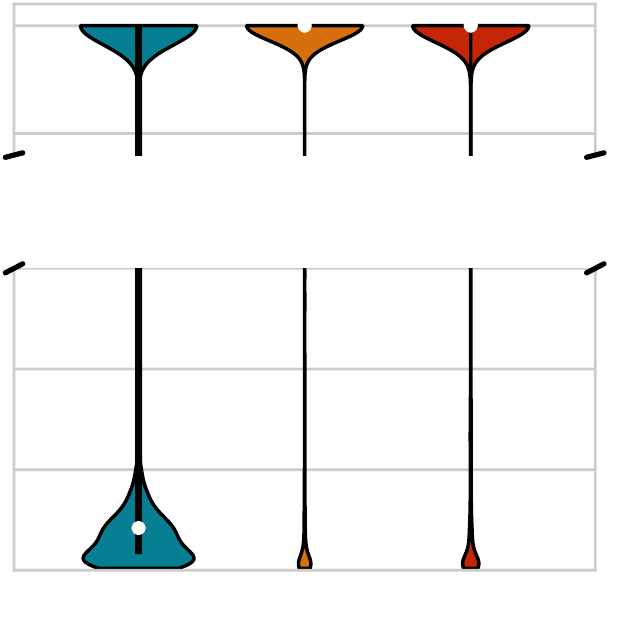} &
         \includegraphics[height=\pich]{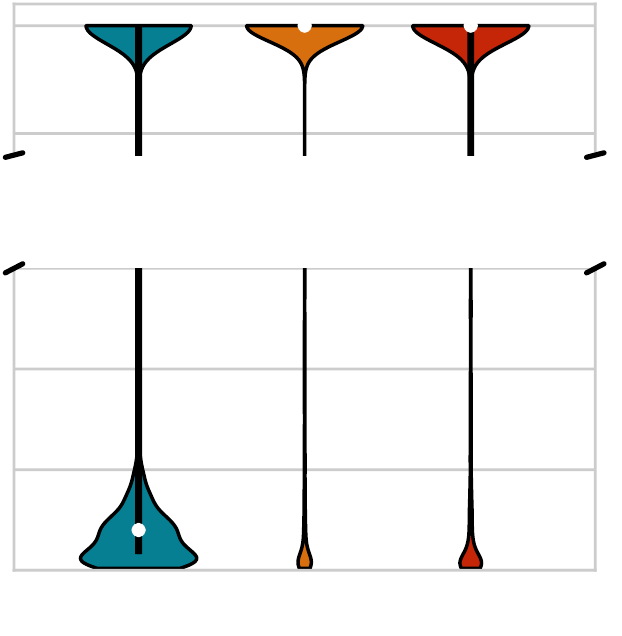} &
         \includegraphics[height=\pich]{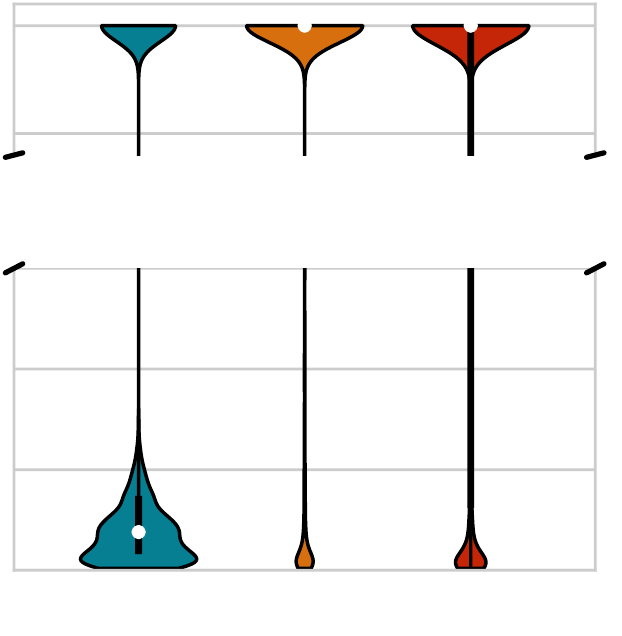} &
         \includegraphics[height=\pich]{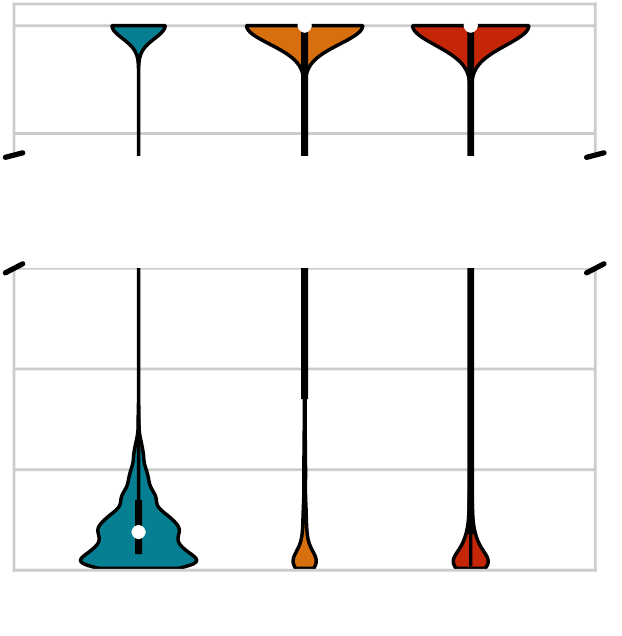} &
         \\
         \textsc{Leaper} &
         \includegraphics[height=\pich]{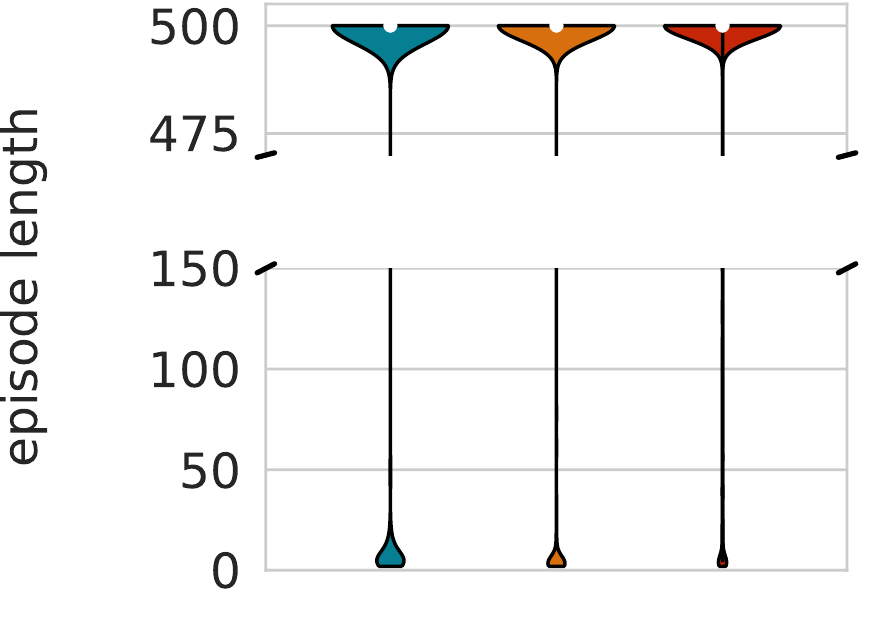} &
         \includegraphics[height=\pich]{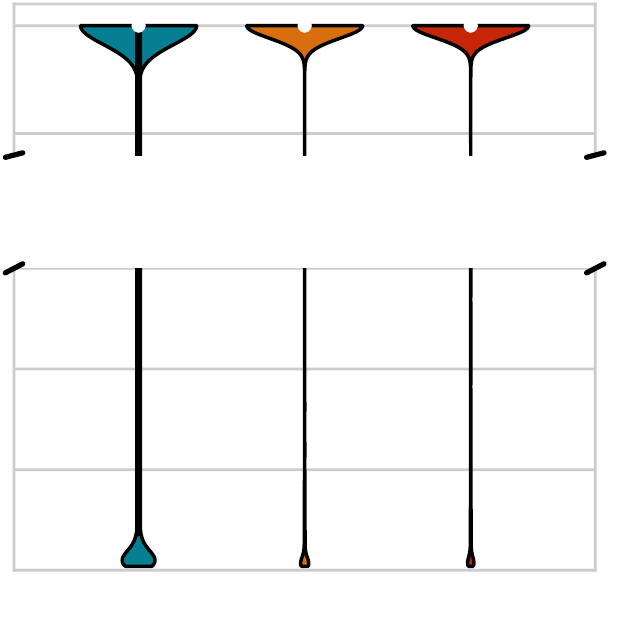} &
         \includegraphics[height=\pich]{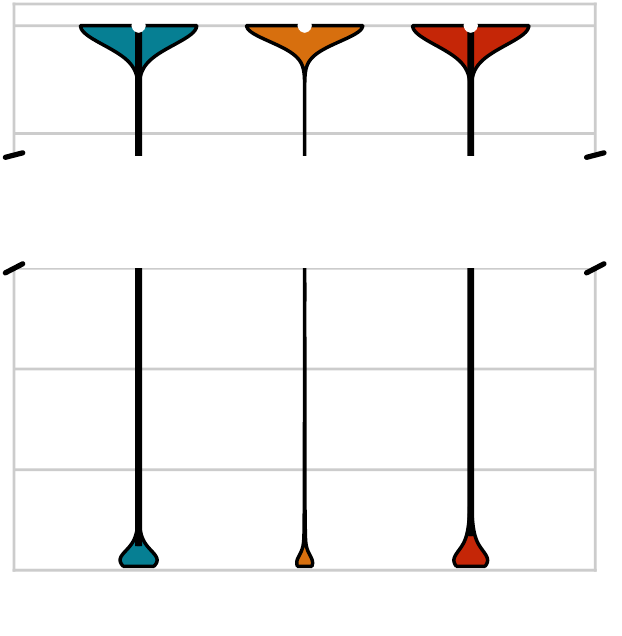} &
         \includegraphics[height=\pich]{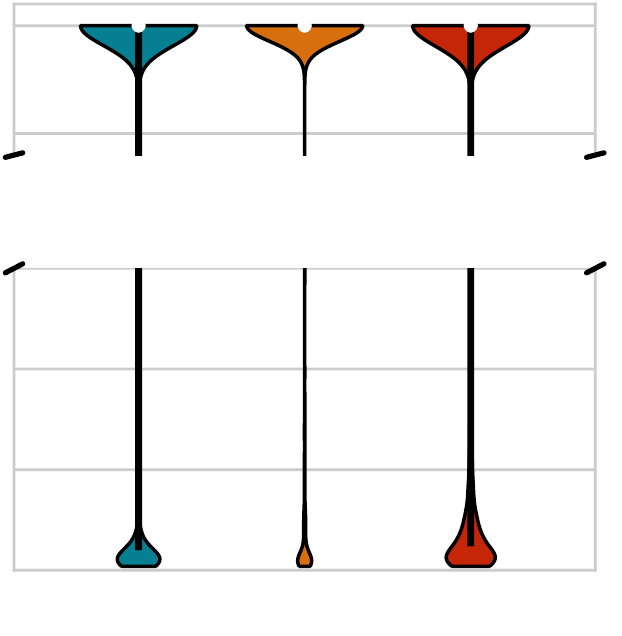} &
         \includegraphics[height=\pich]{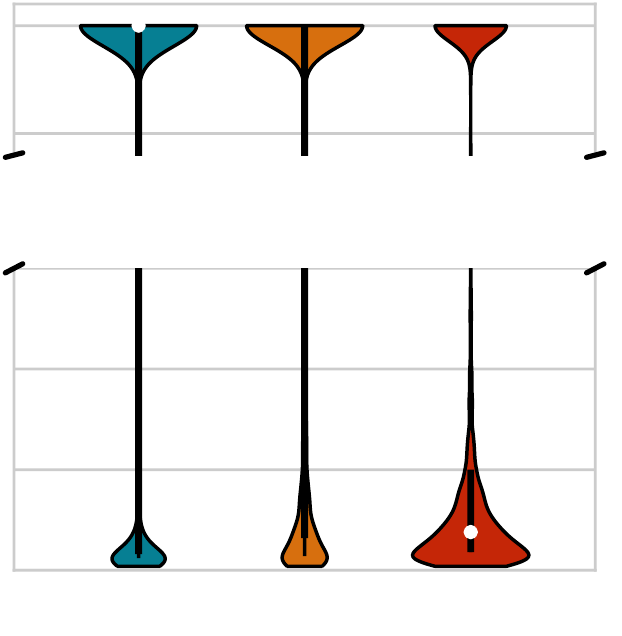} &
         \includegraphics[height=\pich]{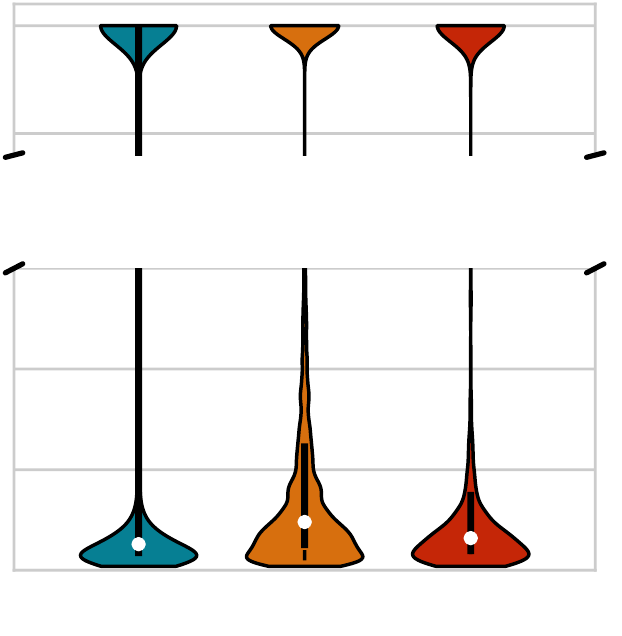} \\

        \multicolumn{8}{c}{
             \includegraphics[width=.3\linewidth]{icml/violin_legend-crop.pdf}
         }
        
    \end{tabularx}
    \vspace{-.7em}
    \caption{Density plots of performance on the test set (1000 unseen levels) after different number of training levels for the \textsc{Maze} and \textsc{Leaper} environments, the white point denotes the median performance on the test set.}
    \label{fig:supp-violin}
\end{figure*}

In addition, \method{} is a general architecture paradigm that may be composed with various different objective functions, including a reinforcement learning formulation.
It would be interesting to see how \method{} behaves when used in such a formulation and if this would lead to even better generalization properties with more data.
We provide training performance curves in Fig. \ref{fig:supp-performance} and density plots for different numbers of training levels in Fig. \ref{fig:supp-violin}.

\section{Data Regularized Actor-Critic}
\label{app:drac-details}

The DrAC algorithm \citep{raileanu2020automatic} attacks the problem of generalization in reinforcement learning from a different angle, namely applying (in some versions optimized) data augmentations to the PPO algorithm.
The main insight is that naively applied data augmentations result in faulty policy gradient estimates because the resulting policy after applying the augmentation is too far from the behavior policy.

To alleviate this, a policy regularization term $G_\pi$ and value function regularization term $G_V$ are added to the PPO objective: 
\begin{equation}
    J_{\text{DrAC}} = J_{\text{PPO}} - \alpha_r ( G_\pi + G_V).
\end{equation}

Furthermore, various augmentations and augmentation selection schemes were proposed.
We ran all of the proposed selectin schemes on our environments, Meta-DrAC, RL2-DrAC, UCB-DrAC and DrAC.
Meta-DrAC meta-learns the weights of a convolutional neural network used for data augmentation.
RL2-DrAC meta-learns a policy that selects an augmentation from a pre-defined set of augmentations.
UCB-DrAC is a bandit formulation of the augmentation selection problem with application of a upper confidence bound algorithm for selection strategy.
DrAC denotes the version with the crop augmentation, which has been shown to work well with more than half of the environments in the ProcGen benchmark.
For more details, we refer the reader to \citet{raileanu2020automatic}.

DrAC's approach to improving generalization is orthogonal to \method{} and the approaches may be composed in order to achieve even better generalization capabilities.

\section{Additional Related Work}
\label{app:additional-references}

\paragraph{Generalization in  reinforcement learning} In addition to the work of \citet{raileanu2020automatic}, there is a plethora of approaches that attempt to improve generalization in reinforcement learning by considering various data augmentation techniques while mainly drawing motivation from supervised learning approaches  \citepsec{kostrikov2020image, laskin2020reinforcement}.
Other approaches combine unsupervised learning with data augmentation \citepsec{srinivas2020curl, zhan2020framework}.

Notably, the problem of sim-to-real transfer can be seen as a problem of generalization to different system dynamics.
Domain randomization \citesec{tobin2017domain}, i.e. augmenting system dynamics in a structured way, has emerged as one of the main techniques for tackling this problem.

\bibliographysec{main}
\bibliographystylesec{apalike}

\end{document}